\documentclass[letterpaper]{article} 
\usepackage{aaai2027}  
\usepackage[hyphens]{url}  
\usepackage{graphicx} 
\graphicspath{{images/}}
\usepackage{natbib}  
\usepackage{caption} 
\usepackage{amsmath}
\usepackage{amssymb}
\usepackage{booktabs}
\usepackage{multirow}


\title{INTERVenE: Temporal-Abstraction-Interval Based Transformers for Short-Horizon Medical Event Prediction}

\author{
    Shahar Oded\textsuperscript{\rm 1},
    Yuval Shahar\textsuperscript{\rm 1}
}

\affiliations{
    \textsuperscript{\rm 1}The Stein Faculty of Computer and Information Science, Ben-Gurion University of the Negev, Beer-Sheva, Israel
}

\begin{document}

\maketitle

\begin{abstract}
Electronic Health Record (EHR) prediction models in the intensive care unit must learn from sparse and irregular measurements while preserving the clinical meaning of time and supporting transparent decision-making. We present INTERVenE, a family of Transformer architectures whose input is an interval-based, knowledge-based temporal abstraction (KBTA), a token stream of named clinical concepts (states, trends, events, contexts) drawn from a curated medical ontology, rather than an unnamed bin index or a raw measurement triplet. This naming layer is what we ask KBTA to do: it makes the model's per-token attributions resolve to clinical concepts by construction. INTERVenE offers two complementary variants: an auto-regressive decoder that generates future abstraction trajectories with a per-step risk readout (localizing \emph{when} and \emph{after which events} risk rises), and a bidirectional encoder for single-pass joint risk and time-to-event prediction. On 57,078 MIMIC-IV admissions, INTERVenE-Enc outperformed GRU-D, STraTS, and KarmaLego, reaching a support-weighted AUPRC of 0.672 ($+0.046$ over the strongest neural baseline, with non-overlapping 95\% bootstrap CIs), AUROC of 0.901, and length-of-stay MAE of 44.4\,h. INTERVenE-Ar complements these predictions with generated abstraction trajectories and evolving risk scores, achieving AUROC of 0.854 and AUPRC of 0.588 under the harder generative readout. An alternative-discretization ablation preserved predictive performance, supporting KBTA's role as a clinically named attribution substrate rather than a prerequisite for accuracy.
\end{abstract}

\section{Introduction}

Short-horizon complication prediction in the intensive care unit (ICU) is challenging because physiological events are embedded in irregular, sparse, and heterogeneous electronic health record streams. Raw observations, such as laboratory measurements, medication administrations, and sporadic interventions, arrive at irregular patient-specific times, and their clinical interpretation often depends on duration, contextual trends, and missingness. Although high-capacity neural models excel on such complex temporal data, their ``black-box'' nature limits clinical trust. This paper asks whether a model that first converts raw observations into \textit{knowledge-based temporal-abstraction} (KBTA) intervals can outperform strong neural and symbolic alternatives while yielding clinically interpretable intermediate representations.

We focus on diabetic ICU admissions in MIMIC-IV \citep{johnson2023mimiciv}. Rather than treating diabetes as a static comorbidity, we recognize that inpatient dysglycemia (blood-glucose disturbances) is a highly dynamic process: rapid fluctuations and persistent trends dictate immediate clinical actions and drive severe downstream complications. Accordingly, our task observes the first 48 hours of each admission and predicts whether six target complications will occur during hours 48--336.

We present INTERVenE, an interval-based Transformer family for temporal EHR data with two variants: \textbf{INTERVenE-Ar}, an autoregressive decoder that forecasts future KBTA abstraction trajectories with token-level risk readouts, and \textbf{INTERVenE-Enc}, a bidirectional encoder for single-pass joint risk and time-to-event prediction. \textbf{INTERVenE-Enc} is probed with sparse transcoders that express token-level attribution scores over named clinical states, trends, and events, while \textbf{INTERVenE-Ar}'s autoregressive decoding provides a direct readout of generated abstraction tokens alongside evolving risk trends. Benchmarked against GRU-D \citep{che2018recurrent}, supervised and self-supervised STraTS \citep{tipirneni2022strats}, and KarmaLego+TPF \citep{moskovitch2015karmalego,sheetrit2019tpf} on 57,078 MIMIC-IV admissions, INTERVenE-Enc reaches a support-weighted AUPRC of 0.672 ($+0.046$ over the strongest neural competitor, with non-overlapping 95\% confidence intervals), while INTERVenE-Ar generates complete admission abstraction trajectories with per-step risk curves and maintains competitive predictive performance. An input-representation ablation demonstrates that KBTA serves as an \emph{interpretability} substrate rather than a predictive prerequisite.

\noindent\textbf{Contributions:}
\begin{itemize}\setlength{\itemsep}{.1em}
  \item A KBTA-based interval tokenization framework casting clinical temporal abstractions as a named-concept Transformer vocabulary.
  \item Two complementary Transformer variants over interval abstractions: INTERVenE-Enc for single-pass risk and time-to-event prediction, and INTERVenE-Ar for autoregressive trajectory generation with per-step risk curves.
  \item A gradient-decoupled sparse-transcoder attribution pipeline whose token-level scores resolve to named clinical concepts by vocabulary construction.
  \item Empirical validation on 57{,}078 MIMIC-IV admissions against four baselines with 2{,}000-resample patient-level bootstrap confidence intervals.
\end{itemize}
\noindent\textbf{Central thesis:} KBTA provides the interpretability substrate; INTERVenE provides the predictive framework operating over it.

\section{Related Work}

\paragraph{Temporal abstraction and interval mining.}
Statistical symbolic representations such as SAX discretize segment averages into symbols \citep{lin2003symbolic}. KBTA instead uses declarative, potentially context-sensitive medical knowledge to represent observations as named clinical states, trends, events, and contexts, such as high glucose or a decreasing glucose trend \citep{shahar1997framework}. KarmaLego mines temporal interval patterns, and Temporal Probabilistic Profiles (TPF) summarize them for prediction \citep{moskovitch2015karmalego,sheetrit2019tpf}; together they form our symbolic baseline. Cumulative stay-time representations encode accumulated time in health states for event-time prediction \citep{katsuki2022cumulative}, whereas INTERVenE retains a sequence of interval boundaries.

\paragraph{Deep learning for irregular clinical sequences.}
GRU-D models missingness and elapsed time through learned decays \citep{che2018recurrent}, while STraTS embeds value--time--variable triplets \citep{tipirneni2022strats}. Other EHR Transformers model longitudinal histories and concept relations \citep{li2020behrt,yang2023transformehr,huang2024heart}. More closely related, Multi-TA combines interval-level clinical-state tokens with temporality-aware BERT learning and temporal augmentation \citep{sohn2024multi}. INTERVenE's specific contribution combines a hierarchy grounded in the KBTA knowledge base with START/END boundary tokens for states, trends, events, and contexts, continuous timing, and complementary encoder and autoregressive objectives.

\paragraph{Explainability in sequential clinical models.}
KBTA represents clinical states and trends as distinct named tokens, making their temporal meaning explicit in the input vocabulary. Sparse-transcoder attribution \citep{dunefsky2024transcoders} assigns signed scores to these tokens, supporting inspection of model sensitivity in clinical terms.

\section{Cohort and Prediction Task}

The cohort contains 57,078 adult diabetes-related admissions from MIMIC-IV. Admissions exceeding 14 days were excluded. For deaths within 30 days after discharge, the RELEASE token was relabeled DEATH, making the terminal categories mutually exclusive and exhaustive in the retained cohort. We focus on diabetic admissions because they provide dense physiological measurements, a mature KBTA ontology, and six clinically meaningful complications with lab-derived onset times, making the cohort well suited to interval-based modeling. Admissions are split at the patient level into 39,954 training, 8,562 validation, and 8,562 held-out test admissions. All methods use the same patient split, 48-hour observation window, and outcome labels.

All models shared a flat vector of demographics and admission-level descriptors. GRU-D and both STraTS variants used irregularly timed laboratory measurements, medication administrations, and interventions; KarmaLego+TPF and INTERVenE used the KBTA states, trends, events, and contexts produced by Mediator \citep{shahar1997framework}. All models used the same 48-hour cutoff. State and trend labels were based on measurements available at interval onset (Appendix~\ref{app:abstraction-construction}).

The prediction window spans hours 48--336 after admission. Target complications occurring during the initial 48-hour observation window are excluded from the positive label, ensuring models predict future events rather than observations already seen. We predict six binary outcomes: death, kidney complication (AKI), hyperglycemia, severe hyperglycemia, hypoglycemia, and severe hypoglycemia. Clinical thresholds are provided in the Appendix. A continuous length-of-stay target serves as an auxiliary regression objective across all models. AUPRC is the primary evaluation metric because the outcomes are imbalanced; AUROC and best F1 are reported as secondary measures. F1 thresholds were selected on the validation set and applied unchanged to the test set. Support-weighted metrics average per-outcome scores weighted by the number of positive cases.

\section{Methods}

\begin{figure*}[t]
    \centering
    \includegraphics[width=\textwidth]{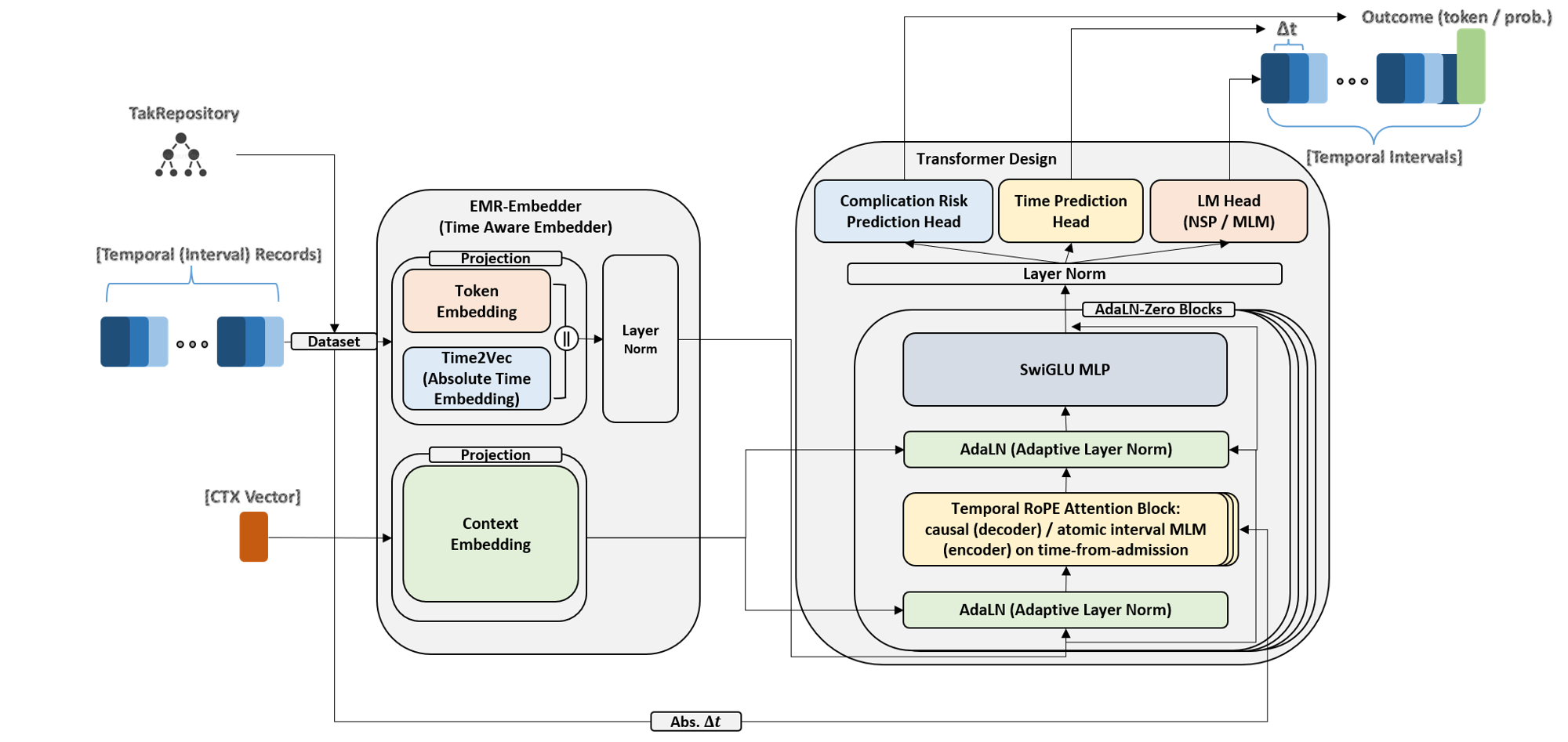}
    \caption{The INTERVenE architecture. Raw clinical sequences are abstracted into KBTA intervals. These intervals are tokenized into a hierarchical vocabulary, enriched with Time2Vec continuous temporal encodings, and processed by either an autoregressive decoder (INTERVenE-Ar) or a bidirectional encoder (INTERVenE-Enc). Temporal Rotary Position Embeddings (RoPE) are applied within the attention layers, while AdaLN-Zero injects static patient context.}
    \label{fig:architecture}
\end{figure*}

\subsection{Shared Representation and Backbone}

The core of INTERVenE is a structured alignment of discrete medical knowledge and continuous temporal dynamics. Raw temporal records are mapped to KBTA intervals using the Mediator temporal abstraction engine \citep{shahar1997framework}, which applies a diabetes-specific knowledge base to identify clinically meaningful states, trends, contexts and events.

To represent variable durations, each interval contributes two timestamped tokens: \emph{START} and \emph{END}. Each boundary combines four hierarchical embeddings (Raw Concept, TAK Abstraction Concept, Concept with Value, and Positional Role) within one vector, without adding sequence positions. For example, a high-glucose interval start maps to \emph{GLUCOSE} $\rightarrow$ \emph{GLUCOSE\_STATE} $\rightarrow$ \emph{GLUCOSE\_STATE\_HIGH} $\rightarrow$ \emph{GLUCOSE\_STATE\_HIGH\_START}. Sharing embedding components encourages related representations: \emph{GLUCOSE\_STATE} and \emph{GLUCOSE\_TREND} share the \emph{GLUCOSE} embedding, while the START and END boundaries of the same interval share three of four embedding components. Figure~\ref{fig:hierarchy-tokenization} illustrates the lookups; learned embedding geometry is shown in Appendix~\ref{app:embedding-geometry}.

To model time, absolute timestamps are anchored to admission and encoded with Time2Vec, which learns linear and periodic temporal components \citep{kazemi2019time2vec}. Unlike standard Transformers, which combine token and positional information through element-wise addition, INTERVenE concatenates the four hierarchical concept embeddings with the Time2Vec representation. The resulting $5d$-dimensional vector is then linearly projected back to the model embedding dimension $d$, preserving concept identity and temporal information as distinct subspaces while allowing the projection layer to learn their interaction. Concretely, each boundary token $c_i$ is represented as

\begin{equation}
\begin{aligned}
z_i =\;&
\Big[
E_1(c_i^{(1)}) \Vert
E_2(c_i^{(2)}) \Vert
E_3(c_i^{(3)}) \Vert\\
&\qquad E_4(c_i^{(4)}) \Vert
\phi_{\mathrm{T2V}}(t_i-t_0)
\Big],\\
h_i^{(0)} =\;& W z_i + b
\end{aligned}
\end{equation}

where $\Vert$ denotes vector concatenation, $E_\ell(c_i^{(\ell)}) \in \mathbb{R}^{d}$ is the embedding of the $\ell$-th hierarchical concept associated with token $i$, and $\phi_{\mathrm{T2V}}(t_i-t_0) \in \mathbb{R}^{d}$ is the Time2Vec encoding of the elapsed time from the admission timestamp $t_0$ to the token timestamp $t_i$. The resulting $5d$-dimensional representation is linearly projected to the model embedding dimension $d$ before entering the Transformer. Temporal rotary position embeddings (RoPE) \citep{su2024roformer} are also applied within the attention layers. Finally, static patient context (e.g., age, sex) is injected directly into each Transformer block through AdaLN-Zero adaptive layer-normalization \citep{peebles2023adaln0}. The raw context attributes are linearly projected to $d_\text{model}$ before entering AdaLN-Zero; this projected vector acts as a global bias to shift the entire event manifold into a patient-specific subspace, utilizing context-dropout during training to regularize the shared latent space.

Training batches group sequences of similar length to reduce padding, with each batch padded to its longest sequence and no fixed token-count cutoff. The encoder processes the observation sequence jointly; autoregressive decoding retains attention history and updates explicit open-interval state.

The embedder, temporal-RoPE attention, and AdaLN-Zero conditioning over SwiGLU blocks form the shared backbone. The variants differ only in attention mask (causal for INTERVenE-Ar, bidirectional for INTERVenE-Enc) and Phase-2/3 objectives; both use the same three-phase schedule. \textbf{Phase 1 (representation)} trains the embedder with a temporal next-token objective: multi-hot BCE over events in a short future window, plus a time-delta regression auxiliary,
\begin{equation}
\mathcal{L}_1 = \mathrm{BCE}_{\text{win}}\big(\hat{y}, y\big) + \lambda_{\Delta t}\,\mathrm{MSE}\big(\widehat{\Delta t}, \Delta t\big)
\end{equation}
The $\Delta t$ term is held behind a BCE-only warmup and unlocked once the main signal is meaningful; its weight $\lambda_{\Delta t}$ is then calibrated once from the loss ratio at unlock and capped at a fixed fraction of the BCE so it never dominates. This phase isolates a patient-aware event representation before the deeper sequence model is trained on top of it.

\subsection{INTERVenE-Ar (Autoregressive Decoder)}

INTERVenE-Ar applies a causal mask and is trained to generate full abstraction trajectories while reading a risk curve at every generated step.

\textbf{Phase 2 (causal pretraining).} The backbone is pretrained with a legality-masked multi-hot next-token BCE under teacher forcing, with clinically illegal continuations removed from the target (e.g., opening a \texttt{GLUCOSE\_STATE\_NORMAL} while \texttt{GLUCOSE\_STATE\_HIGH} is still active; rules in Appendix~\ref{app:legality-masking}). In irregular clinical sequences, token adjacency does not reflect elapsed time. We therefore use time-decayed targets over future occurrences across the full vocabulary, emphasizing nearer events and providing anticipatory supervision beyond the next token, particularly for sparse terminal events. For each query position $t$ and token class $v$, the soft target is

\begin{equation}
  \tilde{y}_v(t) = \begin{cases}
    \exp(-\Delta t_{tv}/\tau_v), & 0<\Delta t_{tv}\le H,\\
    0, & \text{otherwise},
  \end{cases}
\end{equation}

where $\Delta t_{tv}$ is the gap to the nearest strictly future occurrence of $v$ ($\infty$ if absent), $H$ is the look-ahead horizon, and $\tau_v$ is a class-specific decay constant. Appendix Table~\ref{tab:abl-ar-all} compares against a short-decay kernel ($\tau=0.5$\,h). Terminal-token decay constants are fixed; outcome-class constants remain learnable (settings in Appendix~\ref{app:implementation}). The outcome head is trained through pairwise ranking of positions with versus without an outcome within $H_{\text{out}}$, as determined by the temporal outcome targets. The combined objective is
\begin{equation}
\begin{aligned}
\mathcal{L}_{\mathrm{Ar},2} ={}& \mathcal{L}_{\mathrm{token}}
+ \lambda_{\mathrm{CE}}\mathcal{L}_{\mathrm{setCE}}
+ \lambda_{\Delta t}\mathcal{L}_{\mathrm{time}}\\
&+ \lambda_{\mathrm{ttt}}\mathcal{L}_{\mathrm{ttt}}
+ \lambda_{\mathrm{rank}}\mathcal{L}_{\mathrm{rank}}.
\end{aligned}
\end{equation}
Here, $\mathcal{L}_{\mathrm{token}}$ is legality-masked, focal-weighted BCE on temporal soft targets. Set cross-entropy concentrates probability on legal target tokens; the timing auxiliary combines a binary time-advance gate with next-timestamp regression; and the time-to-terminal auxiliary regresses $\log(1+\delta_i)$, where $\delta_i$ is hours to the next terminal event. These three auxiliaries activate after token-only warmup, followed by outcome ranking when validation loss plateaus. Weights are calibrated once relative to the token loss; precise definitions and scheduling appear in Appendix~\ref{app:ar-phase2-loss}. Curriculum-by-masking (CBM) atomically corrupts linked START/END pairs on the input side to improve tolerance to corrupted context.

\textbf{Phase 3 (risk-head alignment).} Phase 3 fine-tunes the risk head under teacher forcing on natural-distribution batches using outcome BCE on time-decayed soft targets together with pairwise ranking, with a reduced backbone learning rate and per-outcome positive weighting to compensate class imbalance. A patient-level attention pool auxiliary trains in parallel: a learnable query per outcome attends over the backbone hidden states to produce a patient-level BCE signal, calibrated once after the first epoch. This phase gave the largest single lift in our ablation (Table~\ref{tab:abl-ar-all}).

\textbf{Inference.} The decoder is seeded with the observation window and generates a free-running trajectory, reading the risk head at each step to produce a per-complication probability curve over time. A \texttt{ttt}-driven \emph{inference gate} uses the model's own time-to-terminal estimate to halt generation, re-calibrating trajectory length post hoc with no retraining. Patients that reach the length cap receive a forced terminal token (DEATH or RELEASE by highest logit), clamped to the generation horizon.

\subsection{INTERVenE-Enc (Bidirectional Encoder)}

INTERVenE-Enc replaces the causal mask with full bidirectional attention, trading trajectory generation for a faster, single-pass discriminative read-out.

\textbf{Phase 2 (masked pretraining).} The backbone is pretrained with masked language modeling using \emph{atomic-interval masking}: roughly 15\% of eligible positions are selected, and linked interval START/END pairs are masked jointly with dedicated \texttt{[MASK\_INTERVAL\_*]} tokens: masking boundaries independently would allow the model to infer a masked boundary from its visible partner, reducing the task to near-trivial position lookup rather than genuine context reconstruction. The combined objective is
\begin{equation}
\mathcal{L}_{\mathrm{Enc},2}
= \mathcal{L}_{\mathrm{MLM}}
+ \lambda_{\mathrm{pos}}\mathcal{L}_{\mathrm{pos}}
+ \lambda_{\mathrm{local}}\mathcal{L}_{\mathrm{local}}.
\end{equation}
Here, $\mathcal{L}_{\mathrm{MLM}}$ is full-vocabulary cross-entropy at masked positions. The temporal auxiliaries are MSE losses: $\mathcal{L}_{\mathrm{pos}}$ predicts normalized time-since-admission at non-padding positions, while $\mathcal{L}_{\mathrm{local}}$ predicts the normalized gap to the nearest eligible unmasked event at masked positions, encouraging temporal placement to remain recoverable when concept identity is hidden. Auxiliary weights are zero during initial MLM-only training and subsequently calibrated relative to the main loss; target definitions and scheduling details appear in Appendix~\ref{app:enc-phase2-loss}.

\textbf{Phase 3 (joint risk and time).} A task module attaches a per-outcome attention pool (the same mechanism as the Ar pool head) feeding a shared MLP and two heads, optimized jointly:
\begin{equation}
\mathcal{L}3 = \mathrm{BCE}_{w}(\text{risk}) + \lambda_{\text{time}}\mathrm{MSE}(\text{time})
\end{equation}
The risk term is a multi-label BCE over the scored outcomes (RELEASE dropped and reported separately as length-of-stay) with per-outcome positive weighting from training prevalence; the time term is a per-outcome MSE in $z$-normalized hours over positive patients only, regressing hours to first occurrence, with the RELEASE slot trained on patients released within the horizon so it doubles as a length-of-stay predictor. The balance weight $\lambda_{\text{time}}{=}0.1$ is the encoder's single dominant lever: shifting toward the risk-dominant regime lifts AUPRC substantially while remaining robust at full-data scale (Appendix Table~\ref{tab:abl-enc-all}). The backbone again runs at a small fraction of the head learning rate.

\textbf{Inference.} A single bidirectional pass yields, per (patient, outcome), a risk probability $P = \sigma(\cdot)$ from the risk head and a time estimate $T = \mathrm{softplus}(\cdot)$ from the time head, with $P_{\text{RELEASE}} = 1 - P(\text{DEATH})$. No trajectory generation or terminal forcing is required, giving markedly lower latency than INTERVenE-Ar.

\subsection{Interpretability via Sparse Transcoders}

Operating on a KBTA vocabulary provides named clinical units for attribution. Sparse transcoders provide a feature basis for probing MLP hidden states beyond attention maps. We probed a trained INTERVenE-Enc with sparse transcoders \citep{dunefsky2024transcoders}. For each encoder block $L$ we fitted a JumpReLU transcoder \citep{rajamanoharan2024jumping} that encodes the block's MLP input $x_L$ (the post-normalisation, pre-MLP activation) into a sparse feature code $f_L$, and reconstructs the MLP output $\hat{y}_L$ from $f_L$ under a reconstruction + L0 objective, reaching high per-layer reconstruction (R$^2\approx 0.97$) with a sparse per-token code. Crucially, the transcoder is never substituted into the network. Instead we attach it via a \emph{gradient-decoupled hook}: the hook injects
\begin{equation}
y^{\mathrm{used}}_L = y^{\mathrm{true}}_L + \big(\hat{y}_L - \mathrm{sg}(\hat{y}_L)\big)
\label{eq:gd-hook}
\end{equation}
where $\mathrm{sg}(\cdot)$ denotes stop-gradient. The added term is numerically zero, so the forward pass is bit-exact to the deployed model; but autograd retains a live path from the risk logit through $\hat{y}_L$ to $f_L$, enabling attribution without altering model behavior. The transcoder supplies a sparse feature basis for attribution; clinical names come from the input vocabulary. The signed attribution score for input token $t$ to outcome $k$ is the layer-summed, first-order input$\times$gradient attribution
\begin{equation}
\mathrm{contrib}_t^{(k)} \;=\; \sum_{L=1}^{N_{\mathrm{layers}}}\,\sum_{j=1}^{d_{\mathrm{feat}}} f_L(t,j)\;\frac{\partial\,\mathrm{risk}_k}{\partial f_L(t,j)},
\label{eq:contrib}
\end{equation}
aggregated by token across patients. Equation~\eqref{eq:contrib} summarizes local sensitivity through the transcoder paths. These scores describe the fitted model rather than patient physiology and do not provide a complete account of an individual prediction. Because every input token is a KBTA abstraction, each $\mathrm{contrib}_t^{(k)}$ resolves to a named clinical concept (a state, trend, or event) rather than a raw numerical artifact.

\subsection{Baselines}

We benchmarked against four baselines spanning the principal modeling paradigms for irregular clinical sequences. \textbf{GRU-D} \citep{che2018recurrent} augments a gated recurrent network with learned exponential decays that model elapsed time between observations and the effects of missing data, providing the canonical recurrent approach to irregular clinical sampling. \textbf{STraTS} \citep{tipirneni2022strats} represents each observation as a (concept, time, value) triplet, embedding continuous values and timestamps before processing the resulting set with a Transformer. We evaluated both its self-supervised variant (forecasting pretraining followed by supervised fine-tuning) and a supervised-only variant trained from scratch, as pretraining did not consistently improve performance on this cohort. \textbf{KarmaLego+TPF} is the only baseline that operates on the same KBTA representation as INTERVenE. KarmaLego \citep{moskovitch2015karmalego} mines frequent Allen-style temporal interval relation patterns (TIRPs) from the interval sequences; we mined patterns on the training split using a class-balanced discovery subset, a reduced three-relation set, 1-minute tolerance, 4-hour maximum gap, and minimum vertical support of 0.20. Temporal Probabilistic Profiles (TPF) \citep{sheetrit2019tpf} summarize each patient by per-pattern count and duration statistics. Features were selected using $\chi^2$/BH-FDR ($q<0.05$), and $\ell_2$-regularized logistic regression classifiers were tuned by grid search on the validation split.

All baselines were re-implemented, trained, and selected using the shared patient-level split, observation window, static context, and validation-based model selection. The comparisons address two axes: KarmaLego+TPF versus INTERVenE-Enc compares symbolic pattern mining with Transformer modeling on the same KBTA interval stream; GRU-D and both STraTS variants versus INTERVenE-Enc compares raw-input neural pipelines with an abstraction-based neural pipeline. The latter jointly varies representation and architecture, including model capacity. GRU-D uses measurement sequences with missingness and elapsed-time inputs, while STraTS uses value--time--variable triplets. Both were evaluated with their native raw-measurement inputs; adapting them to interval tokens would constitute a separate baseline configuration.

\section{Results}

Table~\ref{tab:headline} summarizes support-weighted headline metrics and Table~\ref{tab:per-outcome} reports the per-outcome breakdown. All methods share the same train/validation/test split and evaluation protocol, with statistical significance assessed by non-overlapping 95\% patient-level bootstrap confidence intervals (2,000 resamples). INTERVenE-Ar is reported for completeness under its generative rollout scoring contract rather than as a like-for-like discriminative baseline.

\begin{table}[t]
  \caption{Headline patient-level performance on the MIMIC-IV diabetic cohort, support-weighted across the six outcomes. AUPRC$_w$ is the primary evaluation metric; AUROC$_w$ and F1$_w$ summarize discrimination; LoS is length-of-stay MAE (hours). Parameter counts correspond to the deployed models; for KarmaLego, the budget is the number of selected sparse TIRP features. Bold indicates the best result in each column.}
  \label{tab:headline}
  \centering
  \resizebox{\columnwidth}{!}{%
  \begin{tabular}{lccccc}
    \toprule
    Method & Params & AUPRC$_w$ (\%) & AUROC$_w$ (\%) & F1$_w$ (\%) & LoS (h) \\
    \midrule
    KarmaLego+TPF   & 25.6k\,feat. & $59.1{\pm}1.2$ & $85.4{\pm}0.5$ & $57.0{\pm}1.0$ & $100.1{\pm}3.4$ \\
    GRU-D           & 47.9k        & $62.1{\pm}1.4$ & $89.0{\pm}0.5$ & $60.2{\pm}1.1$ & $47.9{\pm}0.8$ \\
    STraTS (pre+ft) & 92.7k        & $61.0{\pm}1.4$ & $89.0{\pm}0.5$ & $59.5{\pm}1.1$ & $48.6{\pm}0.9$ \\
    ss-STraTS       & 86.5k        & $62.7{\pm}1.4$ & $89.5{\pm}0.5$ & $60.8{\pm}1.1$ & $48.5{\pm}0.9$ \\
    INTERVenE-Ar    & 1.77M        & $58.8{\pm}1.3$ & $85.4{\pm}0.7$ & $57.2{\pm}1.1$ & $65.1{\pm}1.1$ \\
    \textbf{INTERVenE-Enc} & 1.85M  & $\mathbf{67.2{\pm}1.2}$ & $\mathbf{90.1{\pm}0.5}$ & $\mathbf{63.1{\pm}1.1}$ & $\mathbf{44.4{\pm}0.9}$ \\
    \bottomrule
  \end{tabular}%
  }

  \smallskip
  \footnotesize INTERVenE-Ar is evaluated from a free-running autoregressive rollout; LoS MAE is read from the terminal generated state. Its headline metrics therefore reflect a generative rollout rather than the single-pass discriminative evaluation used by the other models and are reported for completeness rather than as a like-for-like comparison. The locked Ar configuration is detailed in Table~\ref{tab:abl-ar-all}.
\end{table}

\begin{table*}[t]
  \caption{Per-outcome AUPRC, AUROC, Best-F1, and time-head MAE. Header rows report the positive prevalence (\emph{prev.}, the AUPRC random baseline) and the median-time baseline MAE (\emph{base MAE}). Bold indicates the best value for each (outcome, metric) pair; MAE is reported only for the INTERVenE variants.}
  \label{tab:per-outcome}
  \centering
  \resizebox{\textwidth}{!}{%
  \begin{tabular}{llcccccc}
    \toprule
    Method & Metric & KIDNEY & Hyperglyc. & Sev.\ Hyper. & Hypoglyc. & Sev.\ Hypo. & DEATH \\
           & \emph{prev.}        & 9.3\%  & 26.4\% & 12.6\% & 5.8\%  & 2.7\%  & 12.9\% \\
           & \emph{base MAE (h)} & 28.8   & 26.6   & 33.2   & 42.6   & 45.3   & 136.9 \\
    \midrule
    \multirow{3}{*}{KarmaLego + TPF}
      & AUPRC (\%) & $70.4 \pm 2.9$ & $77.6 \pm 1.4$ & $52.1 \pm 2.9$ & $21.1 \pm 3.1$ & $8.1 \pm 2.1$ & $47.7 \pm 2.9$ \\
      & AUROC (\%) & $91.0 \pm 1.2$ & $88.4 \pm 0.7$ & $84.7 \pm 1.2$ & $75.4 \pm 2.1$ & $73.4 \pm 3.0$ & $82.7 \pm 1.2$ \\
      & Best-F1 (\%) & $67.4 \pm 2.5$ & $69.5 \pm 1.4$ & $53.2 \pm 2.1$ & $28.0 \pm 3.1$ & $15.5 \pm 2.9$ & $49.2 \pm 2.2$ \\
    \midrule
    \multirow{3}{*}{GRU-D}
      & AUPRC (\%) & $72.9 \pm 2.9$ & $77.8 \pm 1.8$ & $55.9 \pm 3.2$ & $24.9 \pm 3.6$ & $12.8 \pm 3.2$ & $55.3 \pm 3.3$ \\
      & AUROC (\%) & $92.1 \pm 1.1$ & $89.9 \pm 0.7$ & $88.9 \pm 1.0$ & $81.7 \pm 1.8$ & $82.6 \pm 2.4$ & $\mathbf{89.6 \pm 1.0}$ \\
      & Best-F1 (\%) & $67.8 \pm 2.7$ & $71.8 \pm 1.5$ & $58.2 \pm 2.2$ & $31.0 \pm 3.2$ & $21.8 \pm 4.4$ & $54.2 \pm 2.5$ \\
    \midrule
    \multirow{3}{*}{STraTS (pre+ft)}
      & AUPRC (\%) & $70.8 \pm 3.2$ & $76.8 \pm 1.8$ & $54.4 \pm 3.3$ & $28.2 \pm 4.0$ & $\mathbf{15.0 \pm 4.0}$ & $52.2 \pm 3.5$ \\
      & AUROC (\%) & $91.7 \pm 1.1$ & $89.6 \pm 0.7$ & $88.8 \pm 1.0$ & $84.6 \pm 1.5$ & $\mathbf{83.9 \pm 2.4}$ & $88.8 \pm 1.0$ \\
      & Best-F1 (\%) & $67.2 \pm 2.7$ & $71.0 \pm 1.4$ & $56.1 \pm 2.2$ & $33.6 \pm 3.2$ & $\mathbf{24.5 \pm 4.7}$ & $52.9 \pm 2.6$ \\
    \midrule
    \multirow{3}{*}{ss-STraTS}
      & AUPRC (\%) & $72.6 \pm 3.2$ & $78.2 \pm 1.7$ & $56.6 \pm 3.1$ & $26.7 \pm 3.6$ & $12.3 \pm 2.8$ & $56.3 \pm 3.3$ \\
      & AUROC (\%) & $92.5 \pm 1.1$ & $90.2 \pm 0.7$ & $89.2 \pm 0.9$ & $\mathbf{84.7 \pm 1.5}$ & $83.3 \pm 2.5$ & $89.5 \pm 1.0$ \\
      & Best-F1 (\%) & $69.5 \pm 2.6$ & $71.8 \pm 1.4$ & $57.5 \pm 2.2$ & $33.6 \pm 3.1$ & $23.0 \pm 4.0$ & $55.3 \pm 2.6$ \\
    \midrule
    \multirow{4}{*}{INTERVenE-Ar}
      & AUPRC (\%) & $70.1 \pm 3.5$ & $77.2 \pm 1.7$ & $58.8 \pm 3.2$ & $27.4 \pm 4.0$ & $12.0 \pm 3.3$ & $36.8 \pm 2.9$ \\
      & AUROC (\%) & $90.6 \pm 1.4$ & $88.4 \pm 0.9$ & $88.4 \pm 1.1$ & $81.4 \pm 2.0$ & $80.4 \pm 2.8$ & $75.2 \pm 1.5$ \\
      & Best-F1 (\%) & $69.5 \pm 2.7$ & $70.2 \pm 1.4$ & $57.6 \pm 2.4$ & $\mathbf{36.5 \pm 3.3}$ & $19.6 \pm 3.8$ & $38.3 \pm 2.3$ \\
      & MAE (h) & $28.0 \pm 3.0$    & $30.6 \pm 1.7$    & $41.4 \pm 2.9$    & $50.6 \pm 4.4$    & $60.4 \pm 7.5$    & $159.6 \pm 10.2$ \\
    \midrule
    \multirow{4}{*}{\textbf{INTERVenE-Enc}}
      & AUPRC (\%) & $\mathbf{79.0 \pm 2.8}$ & $\mathbf{81.8 \pm 1.4}$ & $\mathbf{62.9 \pm 3.2}$ & $\mathbf{30.8 \pm 4.2}$ & $14.4 \pm 3.6$ & $\mathbf{60.5 \pm 3.0}$ \\
      & AUROC (\%) & $\mathbf{94.4 \pm 0.9}$ & $\mathbf{91.1 \pm 0.7}$ & $\mathbf{90.1 \pm 1.0}$ & $84.0 \pm 1.7$ & $\mathbf{83.9 \pm 2.4}$ & $88.8 \pm 1.0$ \\
      & Best-F1 (\%) & $\mathbf{73.7 \pm 2.5}$ & $\mathbf{73.3 \pm 1.4}$ & $\mathbf{60.3 \pm 2.4}$ & $36.3 \pm 3.5$ & $23.4 \pm 4.2$ & $\mathbf{57.6 \pm 2.3}$ \\
      & MAE (h) & $18.4 \pm 2.4$    & $18.5 \pm 1.1$    & $25.7 \pm 2.0$    & $34.1 \pm 3.1$    & $38.8 \pm 4.7$    & $123.0 \pm 8.1$ \\
    \bottomrule
  \end{tabular}%
  }

  \smallskip
  \footnotesize Time MAEs are reported only for the INTERVenE variants because the baselines lack per-outcome time heads. INTERVenE-Ar's MAE corresponds to the peak predicted probability along the generated trajectory matched to the nearest ground-truth event.
\end{table*}

INTERVenE-Enc achieves the strongest overall performance, improving the primary metric to a support-weighted AUPRC of $0.672$ ($+0.046$ over the strongest neural baseline, ss-STraTS, and $+0.081$ over KarmaLego; non-overlapping 95\% CIs). It also achieves the best AUROC$_w$, Best-F1$_w$, and LoS MAE, making it the strongest single-pass discriminative model in the comparison. INTERVenE-Ar attains competitive performance despite the substantially harder generative evaluation protocol, in which predictions are scored only after a full autoregressive rollout. Its strongest performance is observed on the dysglycemia outcomes (Appendix Fig.~\ref{fig:ar-trajectory}), while DEATH remains challenging because of the long prediction horizon and sparse terminal events.

Per-outcome (Table~\ref{tab:per-outcome}), INTERVenE-Enc achieves the highest AUPRC on five of the six outcomes, with the largest improvements for KIDNEY ($0.790$ vs.\ $0.726$) and DEATH ($0.605$ vs.\ $0.563$). STraTS has the highest AUPRC point estimate for the rarest outcome (SEVERE HYPOGLYCEMIA; $2.7$\% prevalence), with overlapping 95\% CIs between STraTS and INTERVenE-Enc. INTERVenE-Enc consistently outperforms the median-time baseline in event-time prediction, reducing MAE by $7$--$14$\,h across all six outcomes.

\paragraph{Calibration.}
Both variants support per-outcome temperature scaling, fitted on the validation split and applied unchanged to the held-out test set. INTERVenE-Enc is well calibrated as emitted (support-weighted ECE $0.016$), with only a small improvement after scaling ($0.013$; Appendix Table~\ref{tab:calibration}). INTERVenE-Ar exhibits higher calibration error under its peak-detector readout ($\approx4$--$12\%$ ECE), but temperature scaling substantially improves the prevalent outcomes (e.g., DEATH: $0.050 \rightarrow 0.011$; Appendix Table~\ref{tab:ar-calibration}).

\paragraph{Interpretability.}
Our exploratory analysis aggregates per-token DEATH attributions across 300 test patients (Figure~\ref{fig:xai}); the complete six-outcome panel is provided in Appendix Figure~\ref{fig:xai-all}. Positive attribution is dominated by \emph{severe hypoalbuminemia}, a well-established mortality marker, followed by elevation of amino-transferase (liver stress) and increasing bicarbonate (acid--base decompensation). Negative attribution highlights recovery-associated concepts, including a decreasing trend of aspartate-aminotransferase and a subcutaneous basal-insulin treatment state, illustrating that attribution distinguishes clinical states from directional trends. The KBTA vocabulary makes these signed scores clinically legible by distinguishing named states and trends. Similar clinically coherent patterns are observed across all six outcomes (Appendix Figure~\ref{fig:xai-all}).

\begin{figure}[t]
  \centering
  \includegraphics[width=\columnwidth]{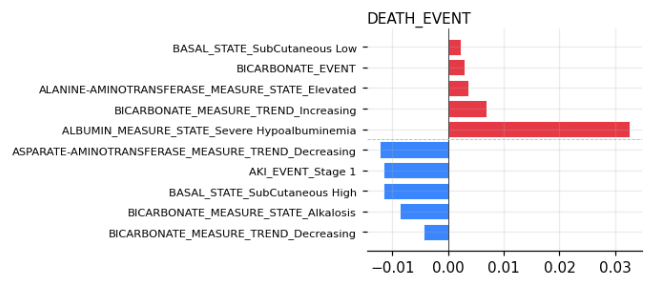}
  \caption{Per-token DEATH attribution scores from gradient-decoupled sparse-transcoder attribution, averaged over 300 test patients. Red and blue denote positive and negative attribution scores, respectively. Full six-outcome panel in Appendix Fig.~\ref{fig:xai-all}.}
  \label{fig:xai}
\end{figure}

\section{Discussion}

These results support two complementary conclusions. First, the AUPRC advantage persisted even in the compact encoder (Appendix Table~\ref{tab:abl-enc-capacity}), indicating that it was not confined to the deployed model size. Representation ablations (Appendix Table~\ref{tab:abl-enc-all}) showed that interval persistence was important for prediction, while replacing KBTA with a simple $\sigma$-based discretization preserved comparable AUPRC and AUROC. Specifically, collapsing all KBTA intervals to a 1-second duration yields a substantial performance drop (weighted AUPRC $0.672 \to 0.593$, LoS MAE $+7.36\,h$), demonstrating that persistence over time of symbolic information is itself an important predictive signal. Second, KBTA contributes primarily through interpretability. By grounding interval tokens in ontology-derived clinical concepts (states, trends, and events), the KBTA vocabulary enables the gradient-decoupled attribution pipeline to produce signed attribution scores over named clinical concepts. Our experiments therefore suggest that KBTA's principal contribution is interpretability, with a modest improvement in length-of-stay prediction, rather than being a prerequisite for predictive accuracy.

The representation helps most where the clinical target is itself an abstraction: severe dysglycemia outcomes are defined by threshold-crossing states and their durations, so the input injects the clinical definition directly into the model. The gain extends to DEATH (AUPRC $0.605$ vs.\ ss-STraTS $0.563$) and length-of-stay ($44.4$\,h vs.\ $\sim$$48$\,h), outcomes not defined by any single KBTA threshold: here the model appears to recognize combinations of abstracted states, including hypoalbuminemia, organ-stress trends, and acid-base shifts, that collectively signal physiological deterioration, consistent with the clinically coherent DEATH drivers in Figure~\ref{fig:xai}.

Finally, the two variants are complementary rather than redundant. \textbf{INTERVenE-Enc} is designed as the production predictor, providing the strongest single-pass accuracy with low-latency inference. \textbf{INTERVenE-Ar} instead prioritizes interpretability, generating complete abstraction trajectories with per-step risk curves that localize \emph{when}, and after \emph{which} events, risk increases---the capability that motivated the autoregressive design. This distinction is reflected in deployment characteristics: INTERVenE-Enc processes patients in $3.3$\,ms each on a single RTX A5000 (batch size 16; $302$ patients/s, allowing the full 8{,}562-patient test set to be evaluated in under 30\,seconds), whereas INTERVenE-Ar requires $255$\,ms per patient ($3.9$ patients/s; CPU inference is impractical), yielding an approximately $77\times$ latency difference and two distinct deployment niches.

\section{Limitations}

The central limitation is intrinsic to the representation. The KBTA layer is only as expressive as its underlying temporal abstraction knowledge (TAK) base: concepts without an explicit abstraction rule receive no symbolic temporal abstractions from Mediator and are therefore absent from INTERVenE's symbolic temporal input. Consequently, extending INTERVenE to new diseases, patient populations, or healthcare settings requires constructing a new domain-specific TAK knowledge base rather than simply retraining on new labels. This is an expert-driven process involving the selection of relevant clinical variables and the definition of abstraction rules, making adaptation substantially more labor-intensive than for end-to-end models trained directly on raw observations. The diabetes knowledge base used here comprises 159 TAK abstraction definitions yielding a vocabulary of 506 named clinical concepts.

A related limitation is granularity. Rule-based abstractions intentionally collapse clinically equivalent values into the same concept; for example, glucose measurements of 260\,mg/dL and 300\,mg/dL both map to the \emph{severe-hyperglycemia} state. The alternative-discretization ablation (Appendix Table~\ref{tab:abl-enc-all}) achieved comparable AUPRC and AUROC using $\sigma$-based bins, suggesting that the proposed architecture is not intrinsically tied to KBTA and can operate over alternative structured discretizations when expert knowledge is unavailable.

Appendix~\ref{app:threshold-proximity} reports threshold proximity and sampled state/trend stability under controlled value perturbations and random observation removal. Beyond the representation, this study evaluates a single MIMIC-IV cohort. External validation, subgroup fairness analyses across demographics and care settings, and attribution robustness across random initializations remain important future directions. Finally, we report patient-level bootstrap confidence intervals rather than multi-seed variance: full training takes approximately 8--12 hours for INTERVenE-Enc and more than 24 hours for INTERVenE-Ar on a single RTX A5000, making multi-seed sweeps prohibitively expensive.

\section{Conclusion}

We presented INTERVenE, a Transformer family for temporal EHR modeling over knowledge-based temporal abstractions. INTERVenE-Enc achieves the strongest predictive performance on the MIMIC-IV diabetic cohort while providing clinically legible, token-level sensitivity scores through gradient-decoupled sparse-transcoder attribution. INTERVenE-Ar complements it with autoregressive abstraction trajectories and per-step risk curves, enabling temporal inspection of how clinical evidence accumulates over an admission. Together, these results demonstrate that competitive predictive performance and attribution over named clinical concepts can be combined within a single neuro-symbolic framework.

\section*{Data and Code Availability}
Source code for INTERVenE, all baselines, and the MIMIC-IV cohort pipeline is included in the supplement; a public repository will be released upon acceptance. Both INTERVenE variants train on a single GPU, while KarmaLego runs CPU-only. Hardware and software details appear in Appendix~\ref{app:implementation}. The pipeline fits within 25\,GB on disk.

\clearpage
\bibliography{bib}

\begin{thebibliography}{17}
\providecommand{\natexlab}[1]{#1}

\bibitem[{Che et~al.(2018)Che, Purushotham, Cho, Sontag, and
  Liu}]{che2018recurrent}
Che, Z.; Purushotham, S.; Cho, K.; Sontag, D.; and Liu, Y. 2018.
\newblock Recurrent Neural Networks for Multivariate Time Series with Missing
  Values.
\newblock \emph{Scientific Reports}, 8(1): 6085.

\bibitem[{Dunefsky, Chlenski, and Nanda(2024)}]{dunefsky2024transcoders}
Dunefsky, J.; Chlenski, P.; and Nanda, N. 2024.
\newblock Transcoders find interpretable llm feature circuits.
\newblock \emph{Advances in Neural Information Processing Systems}, 37:
  24375--24410.

\bibitem[{Huang et~al.(2024)Huang, Rizvi, Thakur, Socrates, Gupta, van Dijk,
  Taylor, and Ying}]{huang2024heart}
Huang, T.; Rizvi, S.~A.; Thakur, R.~K.; Socrates, V.; Gupta, M.; van Dijk, D.;
  Taylor, R.~A.; and Ying, R. 2024.
\newblock HEART: Learning better representation of EHR data with a
  heterogeneous relation-aware transformer.
\newblock \emph{Journal of Biomedical Informatics}, 159: 104741.

\bibitem[{Johnson et~al.(2023)Johnson, Bulgarelli, Shen, Gayles, Shammout,
  Horng, Pollard, Hao, Moody, Gow, Lehman, Celi, and Mark}]{johnson2023mimiciv}
Johnson, A. E.~W.; Bulgarelli, L.; Shen, L.; Gayles, A.; Shammout, A.; Horng,
  S.; Pollard, T.~J.; Hao, S.; Moody, B.; Gow, B.; Lehman, L.-w.~H.; Celi,
  L.~A.; and Mark, R.~G. 2023.
\newblock {MIMIC-IV}, a Freely Accessible Electronic Health Record Dataset.
\newblock \emph{Scientific Data}, 10(1): 1.

\bibitem[{Katsuki et~al.(2022)Katsuki, Miyaguchi, Koseki, Iwamori, Yanagiya,
  and Suzuki}]{katsuki2022cumulative}
Katsuki, T.; Miyaguchi, K.; Koseki, A.; Iwamori, T.; Yanagiya, R.; and Suzuki,
  A. 2022.
\newblock Cumulative stay-time representation for electronic health records in
  medical event time prediction.
\newblock \emph{arXiv preprint arXiv:2204.13451}.

\bibitem[{Kazemi et~al.(2019)Kazemi, Goel, Eghbali, Ramanan, Sahota, Thakur,
  Wu, Smyth, Poupart, and Brubaker}]{kazemi2019time2vec}
Kazemi, S.~M.; Goel, R.; Eghbali, S.; Ramanan, J.; Sahota, J.; Thakur, S.; Wu,
  S.; Smyth, C.; Poupart, P.; and Brubaker, M. 2019.
\newblock {Time2Vec}: Learning a Vector Representation of Time.
\newblock arXiv:1907.05321.

\bibitem[{Li et~al.(2020)Li, Rao, Solares, Hassaine, Ramakrishnan, Canoy, Zhu,
  Rahimi, and Salimi-Khorshidi}]{li2020behrt}
Li, Y.; Rao, S.; Solares, J. R.~A.; Hassaine, A.; Ramakrishnan, R.; Canoy, D.;
  Zhu, Y.; Rahimi, K.; and Salimi-Khorshidi, G. 2020.
\newblock {BEHRT}: Transformer for Electronic Health Records.
\newblock \emph{Scientific Reports}, 10(1): 7155.

\bibitem[{Lin et~al.(2003)Lin, Keogh, Lonardi, and Chiu}]{lin2003symbolic}
Lin, J.; Keogh, E.; Lonardi, S.; and Chiu, B. 2003.
\newblock A symbolic representation of time series, with implications for
  streaming algorithms.
\newblock In \emph{Proceedings of the 8th ACM SIGMOD workshop on Research
  issues in data mining and knowledge discovery}, 2--11.

\bibitem[{Moskovitch and Shahar(2015)}]{moskovitch2015karmalego}
Moskovitch, R.; and Shahar, Y. 2015.
\newblock Classification of Multivariate Time Series via Temporal Abstraction
  and Time Intervals Mining.
\newblock \emph{Knowledge and Information Systems}, 45(1): 35--74.

\bibitem[{Peebles and Xie(2023)}]{peebles2023adaln0}
Peebles, W.; and Xie, S. 2023.
\newblock Scalable Diffusion Models with Transformers.
\newblock In \emph{Proceedings of the IEEE/CVF International Conference on
  Computer Vision}, 4195--4205.

\bibitem[{Rajamanoharan et~al.(2024)Rajamanoharan, Lieberum, Sonnerat, Conmy,
  Varma, Kram{\'a}r, and Nanda}]{rajamanoharan2024jumping}
Rajamanoharan, S.; Lieberum, T.; Sonnerat, N.; Conmy, A.; Varma, V.;
  Kram{\'a}r, J.; and Nanda, N. 2024.
\newblock Jumping ahead: Improving reconstruction fidelity with jumprelu sparse
  autoencoders.
\newblock \emph{arXiv preprint arXiv:2407.14435}.

\bibitem[{Shahar(1997)}]{shahar1997framework}
Shahar, Y. 1997.
\newblock A framework for knowledge-based temporal abstraction.
\newblock \emph{Artificial intelligence}, 90(1-2): 79--133.

\bibitem[{Sheetrit et~al.(2019)Sheetrit, Nissim, Klimov, and
  Shahar}]{sheetrit2019tpf}
Sheetrit, E.; Nissim, N.; Klimov, D.; and Shahar, Y. 2019.
\newblock Temporal Probabilistic Profiles for Sepsis Prediction in the {ICU}.
\newblock In \emph{Proceedings of the 25th ACM SIGKDD International Conference
  on Knowledge Discovery and Data Mining}, 2961--2969.

\bibitem[{Sohn et~al.(2024)Sohn, Park, Park, and Chi}]{sohn2024multi}
Sohn, H.; Park, K.; Park, B.; and Chi, M. 2024.
\newblock {Multi-TA}: Multilevel Temporal Augmentation for Robust Septic Shock
  Early Prediction.
\newblock In \emph{IJCAI}, 6035--6043.

\bibitem[{Su et~al.(2024)Su, Ahmed, Lu, Pan, Bo, and Liu}]{su2024roformer}
Su, J.; Ahmed, M.; Lu, Y.; Pan, S.; Bo, W.; and Liu, Y. 2024.
\newblock {RoFormer}: Enhanced Transformer with Rotary Position Embedding.
\newblock \emph{Neurocomputing}, 568: 127063.

\bibitem[{Tipirneni and Reddy(2022)}]{tipirneni2022strats}
Tipirneni, S.; and Reddy, C.~K. 2022.
\newblock Self-Supervised Transformer for Sparse and Irregularly Sampled
  Multivariate Clinical Time-Series.
\newblock \emph{ACM Transactions on Knowledge Discovery from Data}, 16(6):
  1--17.

\bibitem[{Yang et~al.(2023)Yang, Mitra, Liu, Berlowitz, and
  Yu}]{yang2023transformehr}
Yang, Z.; Mitra, A.; Liu, W.; Berlowitz, D.; and Yu, H. 2023.
\newblock {TransformEHR}: Transformer-Based Encoder-Decoder Generative Model to
  Enhance Prediction of Disease Outcomes Using Electronic Health Records.
\newblock \emph{Nature Communications}, 14(1): 7857.

\end{thebibliography}

\appendix
\setcounter{secnumdepth}{2}

\setcounter{topnumber}{3}
\setcounter{bottomnumber}{2}
\setcounter{totalnumber}{5}
\setcounter{dbltopnumber}{3}
\renewcommand{\topfraction}{0.9}
\renewcommand{\bottomfraction}{0.8}
\renewcommand{\textfraction}{0.1}
\renewcommand{\floatpagefraction}{0.8}
\renewcommand{\dbltopfraction}{0.9}
\renewcommand{\dblfloatpagefraction}{0.8}
\makeatletter
\setlength{\@fptop}{0pt}
\setlength{\@fpsep}{12pt}
\setlength{\@fpbot}{0pt plus 1fil}
\setlength{\@dblfptop}{0pt}
\setlength{\@dblfpsep}{12pt}
\setlength{\@dblfpbot}{0pt plus 1fil}
\makeatother

\begin{center}
  \Large\textbf{Appendix}
\end{center}

\section{Clinical Thresholds and Input Abstraction Rules}
\label{app:clinical-thresholds}

To obtain precise event times, all target complications are defined directly from physiological measurements rather than diagnosis codes. The six outcomes predicted during the 48--336\,h prediction window are defined by the following clinical thresholds applied directly to raw MIMIC-IV laboratory values:

\begin{itemize}
    \item \textbf{Hyperglycemia:} A glucose measurement $\ge$ 250 mg/dL, OR a glucose measurement $\ge$ 180 mg/dL with the immediately preceding glucose measurement also $\ge$ 180 mg/dL, no more than 24 hours earlier (recurrent).
    \item \textbf{Severe Hyperglycemia:} A single glucose measurement $\ge$ 250 mg/dL.
    \item \textbf{Hypoglycemia:} A glucose measurement $\le$ 54 mg/dL, OR a glucose measurement $\le$ 70 mg/dL with the immediately preceding measurement between 20 and 70 mg/dL, no more than 24 hours earlier (recurrent).
    \item \textbf{Severe Hypoglycemia:} A single glucose measurement $\le$ 54 mg/dL.
    \item \textbf{Kidney Complication (Acute Kidney Injury):} A serum creatinine measurement $\ge$ 4.0 mg/dL (AKI Stage 3 criterion), OR a serum creatinine measurement that is $\ge$ 2.0 times the baseline creatinine measurement recorded at admission (AKI Stage 2+ criterion).
    \item \textbf{Death:} Defined by the patient's recorded time of death. To capture all relevant critical care mortality, this includes hospital mortality as well as 30-day post-discharge mortality where linked via state death records.
\end{itemize}

By deriving outcome labels from timestamped physiological measurements whenever possible, the prediction task requires forecasting clinically meaningful deterioration rather than recovering broad administrative diagnoses.

\paragraph{State and trend interval construction.}
\label{app:abstraction-construction}
TAK rules assigned state labels to observed values at measurement time. Observations with the same state label were merged under the configured persistence and interpolation rules; state endpoints were limited by persistence after the last merged measurement and the next measurement time.

For trends, at measurement time $t_i$, an OLS slope $s_i$ was fitted to observations in the trailing window $[t_i-w,t_i]$. For the TAK significant-variation threshold $\delta$, the label was Increasing when $s_iw\ge\delta$, Decreasing when $s_iw\le-\delta$, and Steady otherwise. Insufficient or degenerate windows were labeled Steady. It applied forward to $[t_i,t_{i+1})$ only when $0<t_{i+1}-t_i\le g$, where $g$ was the TAK maximum gap. Larger gaps were omitted; the final point produced no interval or artificial duration. Contiguous intervals with identical labels were merged. In both cases, labels were determined at the starting measurement; subsequent observations governed persistence or termination.

\subsection{Threshold Proximity and Abstraction Stability}
\label{app:threshold-proximity}

\begin{figure*}[!t]
  \centering
  \includegraphics[width=\textwidth,height=0.46\textheight,keepaspectratio]{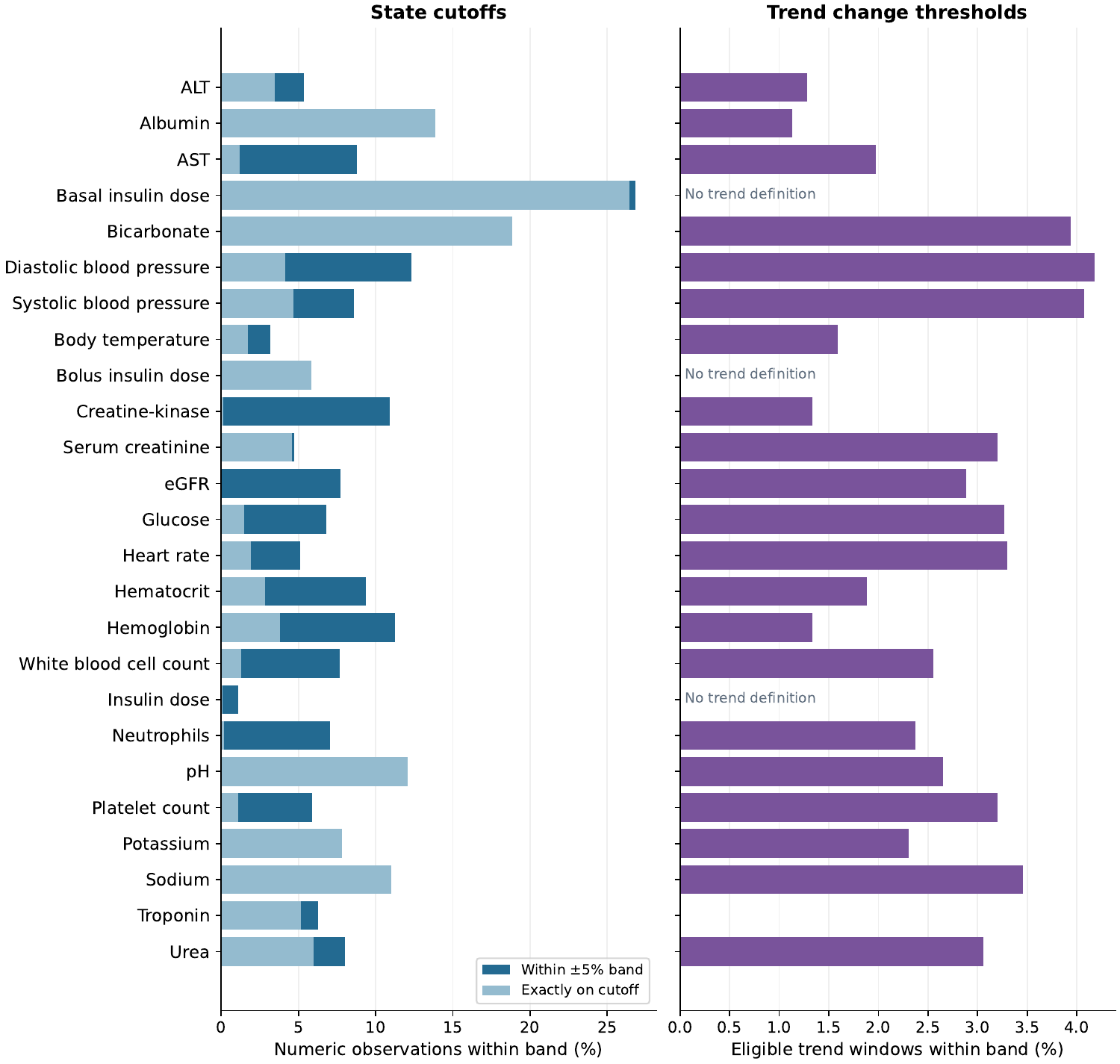}
  \caption{Proximity to TAK boundaries in the first 48 hours. Left: numeric observations within bands of $\pm5\%$ of local state-cutoff spacing; lighter bars show the subset exactly on a cutoff. Right: eligible regression windows within $\pm5\%$ of the trend-change threshold around either signed boundary. Windows with insufficient distinct timestamps were excluded from the trend denominator. Three insulin-dose concepts had no trend definition.}
  \label{fig:threshold-proximity}
\end{figure*}

Knowledge-based state thresholds encoded clinically meaningful value ranges, while trends thresholded regression-estimated change over TAK-specific lookback windows. We assessed proximity to these fixed boundaries in the first 48 hours of all 57,078 admissions, using the TAK repository to resolve direct numeric inputs and excluding derived numeric concepts. The analysis covered 9.43 million observations across 25 state concepts and 7.97 million eligible windows across 22 trend concepts. State bands had half-widths of 5\% of the smaller adjacent cutoff spacing (the nearest spacing at outer cutoffs); trend bands used 5\% of the significant-variation threshold around either signed change boundary. For glucose, these corresponded to 178--182\,mg/dL around the state cutoff of 180, and estimated changes of 38--42 or $-42$--$-38$ around trend thresholds of $\pm40$. Overall, 8.34\% of state observations and 3.45\% of trend windows fell within these bands; 3.98\% of state observations matched a cutoff exactly (Figure~\ref{fig:threshold-proximity}).

\paragraph{Sensitivity to perturbed values and missing observations.}
We reran the unchanged Mediator state/trend implementations on 500 admissions sampled across joint quartiles of first-48-hour observation count and recorded terminal-event span. Numeric inputs received uniform perturbations bounded by $\pm1\%$ or $\pm5\%$ of local spacing at the nearest state cutoff, or independent removal with probability 5\% or 10\%, with two random realizations per condition. Interval agreement was matching-label duration divided by the union of baseline and altered coverage, accounting for label changes and lost or added coverage. Results were weighted by sampling strata; 95\% intervals used 2,000 admission-level bootstrap resamples (Table~\ref{tab:abstraction-stability}). Agreement remained high under the tested value perturbations; trends were more sensitive to random observation removal.

\begin{table}[!ht]
  \centering
  \small
  \setlength{\tabcolsep}{3pt}
  \caption{Sampled abstraction stability. Interval agreement (\%) with 95\% bootstrap intervals. Perturbation bounds are relative to local state-cutoff spacing.}
  \label{tab:abstraction-stability}
  \begin{tabular}{lcc}
    \toprule
    Condition & States & Trends \\
    \midrule
    Perturb $\pm1\%$ & 97.7 [97.5, 97.9] & 99.6 [99.5, 99.6] \\
    Perturb $\pm5\%$ & 97.2 [97.0, 97.4] & 97.9 [97.8, 98.1] \\
    Remove 5\% & 97.2 [97.1, 97.4] & 92.8 [92.5, 93.1] \\
    Remove 10\% & 93.9 [93.7, 94.2] & 85.6 [85.2, 86.1] \\
    \bottomrule
  \end{tabular}
\end{table}

\section{Architectural Ablation Trail}
\label{app:ablation-trail}

The architectures reported in the main paper are the endpoints of an iterative design search documented here. Early experiments were conducted on a 10k-admission design subset to explore architectural choices, followed by confirmation on the full training cohort using the final recipes. Intermediate design stages report only the metrics tracked during the corresponding search and therefore contain ``--'' where metrics were not evaluated.

\subsection{Decoder ablations}

\begin{table*}[!t]
\caption{INTERVenE-Ar architectural trail, stratified by sample size. Reading order: foundational 10k build-up; 10k component-drop ablations of the locked recipe; full-cohort build-up to the deployed recipe. \texttt{-} = not measured at that design stage. Intermediate rows reflect the configuration \emph{as it stood} at each design point, not the deployed recipe with one component removed. Full-cohort: 95\% bootstrap CI half-widths (2{,}000-resample); 10k: point estimates. \emph{gen/GT}: median generated/ground-truth length ratio.}
\label{tab:abl-ar-all}
\centering
\small
\setlength{\tabcolsep}{4pt}
\begin{tabular}{llcccc}
\toprule
Recipe arm & Sample & AUROC$_w$ (\%) & AUPRC$_w$ (\%) & LoS MAE (h) & gen/GT \\
\midrule
Baseline decoder (no TTT aux / pool / gate)      & 10k  & $83.6$                    & $53.9$                    & -                       & $0.60$ \\
\quad + TTT aux head + TTT-guided inference      & 10k  & $84.0$                    & $54.2$                    & -                       & $0.72$ \\
\midrule
\textbf{Locked recipe (10k retrain, anchor)} $=$ \textbf{above} $+$ \textbf{patient-pool aux}     & 10k  & $\mathbf{84.5}$           & $\mathbf{56.2}$           & $71.3$                  & $0.44$ \\
\quad $-$ Phase-3 finetune (eval-only on P2)     & 10k  & $54.7$                    & $21.3$                    & $67.8$                  & $0.66$ \\
\quad $-$ RoPE                                   & 10k  & $83.0$                    & $52.5$                    & $98.3$                  & $1.79$ \\
\quad $-$ CBM                                    & 10k  & $83.4$                    & $53.4$                    & $71.0$                  & $1.09$ \\
\quad Short-decay kernel ($\tau{=}0.5$\,h)          & 10k  & $80.8$                    & $50.0$                    & $111.7$                 & $2.16$ \\
\quad DD damping $\sqrt{}$                       & 10k  & $80.1$                    & $49.3$                    & $69.6$                  & $1.16$ \\
\quad DD damping $\log$ (under-correct)          & 10k  & $79.7$                    & $50.3$                    & $95.7$                  & $1.79$ \\
\midrule
Prior canonical (ttt cap $0.30$, no DD weighting)       & full & $83.2{\pm}0.7$          & $51.9{\pm}1.5$          & $147.7{\pm}1.9$         & $2.82$ \\
\quad + ttt cap $0.30{\to}0.60$            & full & $83.2{\pm}0.7$          & $56.2{\pm}1.3$          & $78.7{\pm}1.0$          & $1.06$ \\
\quad + DD-$p^{0.33}$ weighting \textbf{= Ar-final (deployed)} & full & $\mathbf{85.4{\pm}0.7}$ & $\mathbf{58.7{\pm}1.3}$ & $\mathbf{65.1{\pm}1.1}$ & $\mathbf{0.90}$ \\
\bottomrule
\end{tabular}
\end{table*}

\textbf{Decoder trail (Table~\ref{tab:abl-ar-all}).}
The 10k design search established the core decoder recipe: the \texttt{ttt} auxiliary head improved trajectory-end anticipation, the patient-pool auxiliary provided the largest discrimination gain, and the \texttt{ttt} inference gate restored realistic trajectory lengths at evaluation time. Component-drop ablations identify Phase-3 fine-tuning as the dominant contributor, while RoPE, CBM, the soft-kernel loss, and the damping function each make measurable contributions to trajectory quality and predictive performance. Full-cohort confirmation shows that the deployed recipe substantially improves discrimination, trajectory extent (gen/GT), and LoS prediction over the prior canonical implementation. \textbf{DD-$p^{0.33}$ weighting} applies the per-terminal-token class-imbalance ratio (neg/pos) raised to the power $0.33$ as the LM-head BCE positive weight. Compared with $\sqrt{\cdot}$ and $\log$ damping, it provides the best compromise between over- and under-generation, restoring realistic trajectory lengths while preserving predictive performance.

\subsection{Encoder ablations}

INTERVenE-Enc builds directly on the decoder backbone, inheriting the architectural components validated in Table~\ref{tab:abl-ar-all} (hierarchical token embeddings, Time2Vec, temporal RoPE, AdaLN-Zero conditioning, the per-outcome attention pool, and the soft-kernel BCE). These components are therefore held fixed; the encoder ablations isolate only the additions specific to the bidirectional setting: Phase-3 risk--time loss balancing, Phase-3 backbone learning, the forward-addition trail of inverse-prevalence weighting, focal loss and CBM masking, and the input-representation alternative.

\begin{table*}[!t]
\caption{INTERVenE-Enc architectural trail, stratified by sample size. Reading order: 10k design search; 10k ablation of the locked recipe; full-cohort build-up to the deployed recipe plus one input-representation alternative. \texttt{-} = not measured at that design stage. CI and intermediate-row conventions as in Table~\ref{tab:abl-ar-all}.}
\label{tab:abl-enc-all}
\centering
\resizebox{\textwidth}{!}{%
\begin{tabular}{llccc}
\toprule
Recipe arm & Sample & AUROC$_w$ (\%) & AUPRC$_w$ (\%) & LoS MAE (h) \\
\midrule
\textbf{Locked recipe (10k retrain, anchor)}             & 10k  & $\mathbf{82.2}$  & $\mathbf{49.7}$  & $53.0 \pm 2.1$ \\
\quad + hierarchical MLM masking                         & 10k  & $82.0$           & $48.2$           & - \\
\quad + backbone LR factor $0.01 \to 0.1$                & 10k  & $81.3$           & $47.1$           & - \\
\quad $-$ Phase-3 backbone learning$^\dagger$            & 10k  & $79.2$           & $42.9$           & $55.3 \pm 2.3$ \\
\midrule
Stripped baseline (full data, KBTA input)                & full & $90.1 \pm 0.5$ & $66.2 \pm 1.3$ & $46.1 \pm 0.9$ \\
\quad + inverse-prevalence pos-weight$^\dagger$          & full & $89.4 \pm 0.5$ & $63.8 \pm 1.3$ & $49.1 \pm 0.8$ \\
\quad + focal loss, $\gamma{=}1.0$                       & full & $90.1 \pm 0.5$ & $66.5 \pm 1.2$ & $44.7 \pm 0.8$ \\
\quad + CBM masking $p{=}0.15$ \textbf{= Enc-final (deployed)} & full & $\mathbf{90.1 \pm 0.5}$ & $\mathbf{67.2 \pm 1.2}$ & $\mathbf{44.4 \pm 0.9}$ \\
\midrule
Enc-final with std-bins input (KBTA replaced)            & full & $90.1 \pm 0.5$ & $67.6 \pm 1.2$ & $47.6 \pm 0.9$ \\
Enc-final with de-intervaled input & full & $86.4 \pm 0.7$ & $59.3 \pm 1.5$ & $51.8 \pm 0.8$ \\
\bottomrule
\end{tabular}%
}

\smallskip
\footnotesize $^\dagger$Non-overlapping bootstrap CIs vs.\ the 10k anchor (backbone freeze) or vs.\ the stripped baseline (inv-prev weighting). Focal loss and std-bins rows are within CI noise on AUPRC/AUROC; std-bins LoS $+3.2$\,h is non-overlapping. Hierarchical MLM masking and backbone-LR rows were discarded after 10k search.
\end{table*}

\textbf{Encoder trail (Table~\ref{tab:abl-enc-all}).}
The 10k design search established the Phase-3 training recipe, showing that reducing the time-loss weight substantially improves discrimination while hierarchical MLM masking provides no measurable benefit. Component-drop ablations identify Phase-3 backbone learning as essential, whereas the full-cohort study shows that inverse-prevalence weighting degrades performance, focal loss has negligible effect, and CBM masking ($p{=}0.15$) provides the only consistent improvement over the stripped baseline. Finally, replacing KBTA with a standard-deviation discretization preserves AUPRC and AUROC but increases LoS error, supporting the use of KBTA for interpretability while retaining a modest regression advantage. Crucially, collapsing the intervals into 1-second point events sharply degrades performance across all metrics (weighted AUPRC dropping to $0.593$), demonstrating that the architecture explicitly relies on capturing continuous temporal durations and state concurrency rather than acting as a simple point-event detector.

\begin{table*}[!t]
\caption{INTERVenE-Enc capacity and head-count ablations (full cohort, locked recipe except the dimension under test). Bold = deployed M-128/h=2 configuration. CI conventions as in Table~\ref{tab:abl-ar-all}; size-sweep arms predated LoS tracking (\texttt{-}).}
\label{tab:abl-enc-capacity}
\centering
\small
\setlength{\tabcolsep}{4pt}
\begin{tabular}{lccccccc}
\toprule
Variant & embed\_dim & n\_head & head\_dim & Params & AUPRC$_w$ (\%) & AUROC$_w$ (\%) & LoS MAE (h) \\
\midrule
M-64                                  & 64   & 1 & 64 & 0.46M & $64.6$                    & $88.9$                    & -                       \\
\textbf{M-128 (Enc-final, deployed)}  & 128  & 2 & 64 & 1.84M & $\mathbf{67.2 \pm 1.2}$ & $\mathbf{90.1 \pm 0.5}$ & $\mathbf{44.4 \pm 0.9}$ \\
M-256                                 & 256  & 4 & 64 & 6.40M & $65.0$                    & $89.5$                    & -                       \\
\midrule
h=8 narrow heads (budget-preserving)  & 128  & 8 & 16 & 1.84M & $65.9 \pm 1.3$          & $90.0 \pm 0.5$          & $46.6 \pm 0.9$          \\
h=8 wide heads (parameter-spending)   & 512  & 8 & 64 & 25.0M & $67.6 \pm 1.2$          & $90.5 \pm 0.5$          & $45.2 \pm 0.9$          \\
\bottomrule
\end{tabular}
\end{table*}

\textbf{Capacity and heads (Table~\ref{tab:abl-enc-capacity}).} Model capacity exhibits a clear inverted-U, with the deployed M-128 configuration providing the best accuracy--efficiency trade-off. Increasing model width or attention-head count beyond this point yields little additional predictive benefit despite substantially larger parameter budgets, while smaller models underfit. The smaller M-64 encoder also exceeded the strongest reported neural baseline, suggesting that the gain was not confined to the deployed model size. All evaluated neural models were modest in absolute size (the deployed INTERVenE-Enc had fewer than two million parameters), but their differing parameter budgets prevent a clean attribution of gains. These comparisons do not isolate representation, architecture, or parameter count.

\textbf{STraTS dimensional shape.} INTERVenE-Enc is not evaluated at STraTS's deployed shape (dim=64, h=16, head\_dim=4, $\sim$86K parameters): M-64 underfits and narrow heads at dim=128 both cost AUPRC and LoS, so combining them would compound both losses. The intended comparison is each model at its own tuned configuration.

\section{Calibration}
\label{app:calibration}

Calibration is evaluated without retraining: locked checkpoints remain unchanged, and only the test-time logit-to-probability mapping is optimized.

\subsection{INTERVenE-Enc}

Per-outcome temperatures $T_k$ are fit by LBFGS on the validation split and applied unchanged to the held-out test split. Rank-based metrics are unaffected by the monotonic scaling.

\begin{table}[!htbp]
  \caption{Enc per-outcome $T_k$ (validation fit) and ECE before/after scaling (test split). $T_k{=}1$ = perfectly calibrated; $T_k{>}1$ = over-confident. Last row: support-weighted aggregate.}
  \label{tab:calibration}
  \centering
  \resizebox{\columnwidth}{!}{%
  \begin{tabular}{lcccc}
    \toprule
    Outcome & $T_k$ & ECE$_\text{uncal}$ & ECE$_\text{cal}$ & $\Delta$ECE \\
    \midrule
    KIDNEY                  & 0.866 & 0.0228 & 0.0151 & $-0.0077$ \\
    SEVERE\_HYPERGLYCEMIA   & 0.907 & 0.0224 & 0.0176 & $-0.0048$ \\
    SEVERE\_HYPOGLYCEMIA    & 0.913 & 0.0101 & 0.0067 & $-0.0033$ \\
    HYPERGLYCEMIA           & 0.970 & 0.0166 & 0.0157 & $-0.0010$ \\
    HYPOGLYCEMIA            & 0.974 & 0.0071 & 0.0080 & $+0.0009$ \\
    DEATH                   & 1.039 & 0.0097 & 0.0075 & $-0.0022$ \\
    \midrule
    Weighted (by $n_\text{pos}$) & --- & $0.0162$ & $0.0134$ & $-0.0027$ \\
    \bottomrule
  \end{tabular}%
  }
\end{table}

\begin{figure*}[!t]
  \centering
  \includegraphics[width=\textwidth]{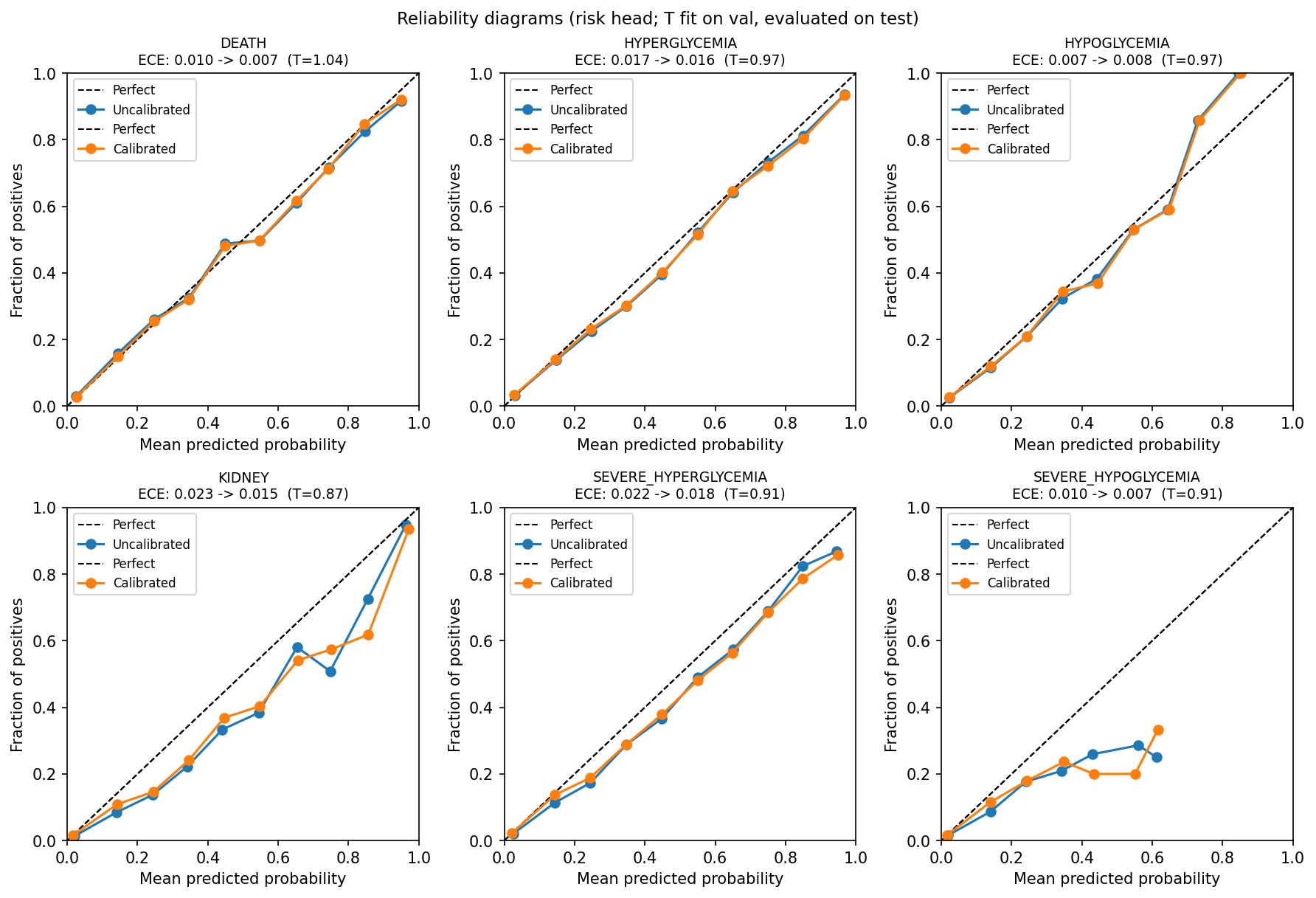}
  \caption{INTERVenE-Enc reliability diagrams (10 equi-width bins, held-out test split). Blue: uncalibrated; orange: after $T_k$ scaling. Dashed diagonal = perfect calibration. Panel titles report ECE and $T_k$.}
  \label{fig:reliability}
\end{figure*}

\textbf{Findings.}
INTERVenE-Enc is well calibrated as emitted, with temperatures close to unity ($T_k\in[0.87,1.04]$). Temperature scaling yields only a modest improvement in support-weighted ECE ($0.016\rightarrow0.013$), consistent with the reliability diagrams (Figure~\ref{fig:reliability}). We nevertheless provide the fitted temperatures alongside the released checkpoint as an optional refinement.

\subsection{INTERVenE-Ar}

Calibration follows the same protocol, using the per-patient maximum outcome probability over the generated trajectory (the same readout used for the headline metrics). Temperatures are fit on the validation split and applied unchanged to the held-out test split.

\begin{table}[!htbp]
  \caption{Ar per-outcome $T_k$ (validation fit) and ECE before/after scaling (test split). $T_k{<}1$ = over-confident; $T_k{>}1$ = under-confident. Readout: per-patient max $P_\text{outcome}$ over generated positions.}
  \label{tab:ar-calibration}
  \centering
  \resizebox{\columnwidth}{!}{%
  \begin{tabular}{lcccc}
    \toprule
    Outcome & $T_k$ & ECE$_\text{uncal}$ & ECE$_\text{cal}$ & $\Delta$ECE \\
    \midrule
    SEVERE\_HYPOGLYCEMIA    & 0.6543 & 0.1119 & 0.0754 & $-0.0365$ \\
    HYPOGLYCEMIA            & 0.7838 & 0.1174 & 0.0911 & $-0.0263$ \\
    KIDNEY                  & 0.9566 & 0.0523 & 0.0490 & $-0.0033$ \\
    SEVERE\_HYPERGLYCEMIA   & 1.0241 & 0.0450 & 0.0469 & $+0.0018$ \\
    HYPERGLYCEMIA           & 1.2365 & 0.0389 & 0.0184 & $-0.0205$ \\
    DEATH                   & 1.4700 & 0.0495 & 0.0108 & $-0.0387$ \\
    \bottomrule
  \end{tabular}%
  }
\end{table}

\begin{figure*}[!t]
  \centering
  \includegraphics[width=\textwidth]{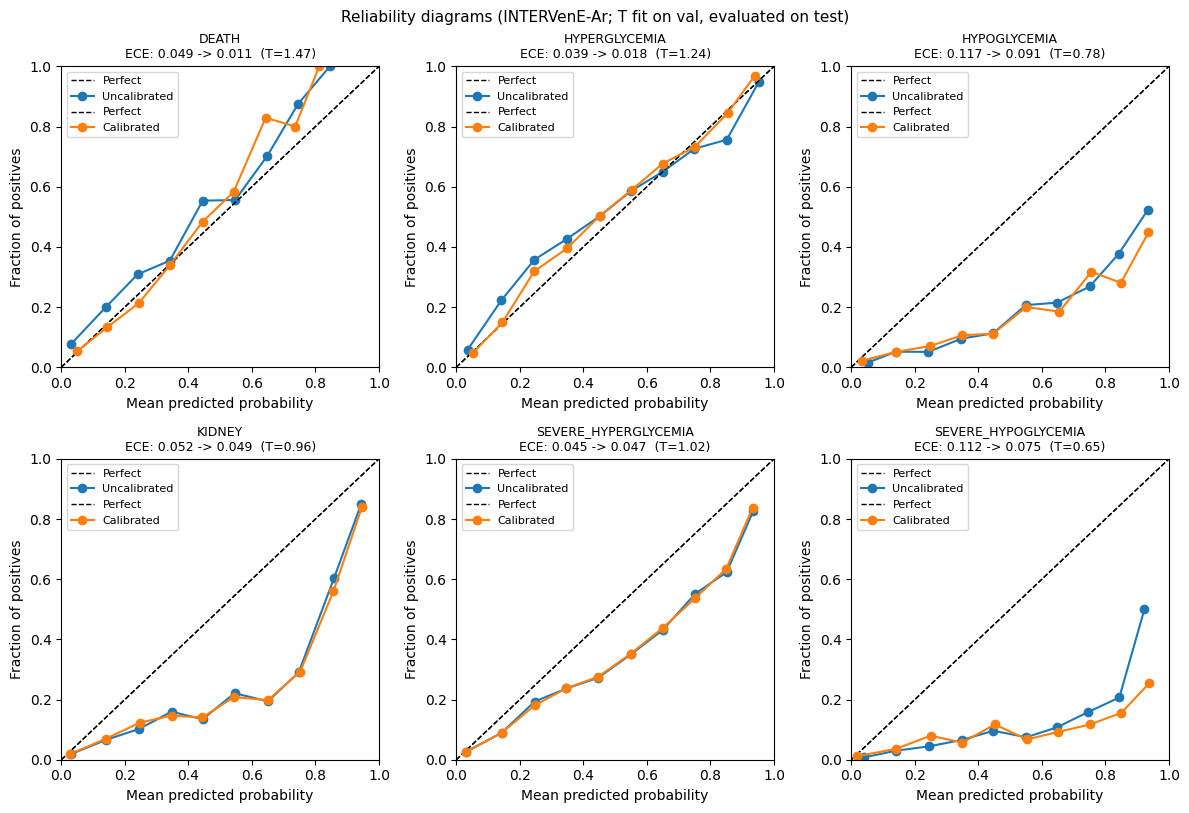}
  \caption{INTERVenE-Ar reliability diagrams (10 equi-width bins, held-out test split). Blue: uncalibrated; orange: after $T_k$ scaling. Prevalent outcomes improve substantially after scaling (DEATH $0.050\!\to\!0.011$); rare hypoglycaemia panels show wider residual deviation consistent with the AUPRC floor in the per-outcome retults table in the main paper.}
  \label{fig:ar-reliability}
\end{figure*}

\textbf{Findings (Ar).}
INTERVenE-Ar exhibits a wider range of fitted temperatures than the encoder ($T_k\in[0.65,1.47]$), reflecting the greater calibration challenge of the autoregressive readout. Temperature scaling substantially improves calibration on the prevalent outcomes (e.g., DEATH: $0.050\rightarrow0.011$ ECE), while the rare hypoglycemia outcomes remain the most difficult to calibrate, consistent with their low prevalence and AUPRC. Reliability diagrams are shown in Figure~\ref{fig:ar-reliability}; fitted temperatures are supplied alongside the checkpoint. Aggregate ECE is omitted because calibration is evaluated on the trajectory-derived peak-probability readout rather than a native discriminative output.

\section{Autoregressive Risk Trajectories}

\begin{figure}[!t]
  \centering
  \includegraphics[width=\columnwidth]{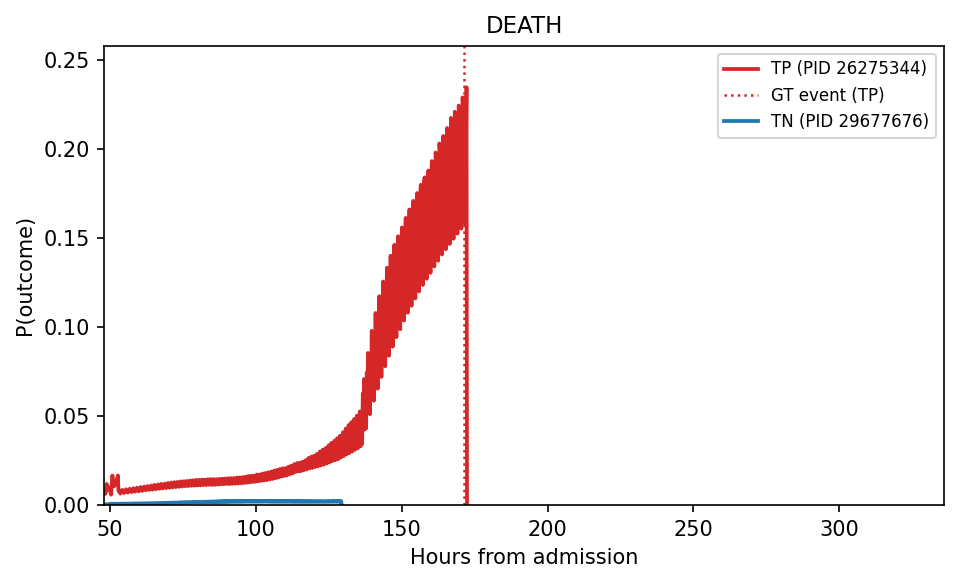}
  \caption{INTERVenE-Ar per-step DEATH risk trajectories for two held-out test patients. Red: a true-positive admission, where risk progressively increases and peaks near the recorded death time ($\approx171$\,h, dotted). Blue: a true-negative admission, where risk remains near baseline throughout the generated trajectory. The $x$-axis begins at the 48\,h observation cutoff; the $y$-axis is normalized to the local peak.}
  \label{fig:ar-trajectory}
\end{figure}

INTERVenE-Ar targets complication forecasting rather than exact reconstruction of the observed future. Irregular sampling and observation gaps can produce different token sequences for clinically similar trajectories. We therefore evaluated complication prediction and generated trajectory extent, reporting the generated-to-ground-truth length ratio in Table~\ref{tab:abl-ar-all}. The training objective emphasizes sparse complication events rather than uniformly weighting trajectory reconstruction; token-level accuracy and broader clinical trajectory quality remain unevaluated.
\label{app:ar-trajectories}
The Ar variant emits a per-step risk curve over the generated trajectory. The risk head is evaluated at every generation step, producing a time-resolved signal that can express multiple risk peaks within an admission. Peak width is governed by the Phase-2 soft-kernel ($\tau=12$\,h for DEATH); absolute risk values should therefore be interpreted relative to $\tau$ rather than as binary decisions. Figure~\ref{fig:ar-trajectory} illustrates two representative test patients.

\section{Implementation Details}
\label{app:implementation}

\begin{figure*}[!t]
  \centering
  \includegraphics[width=\textwidth]{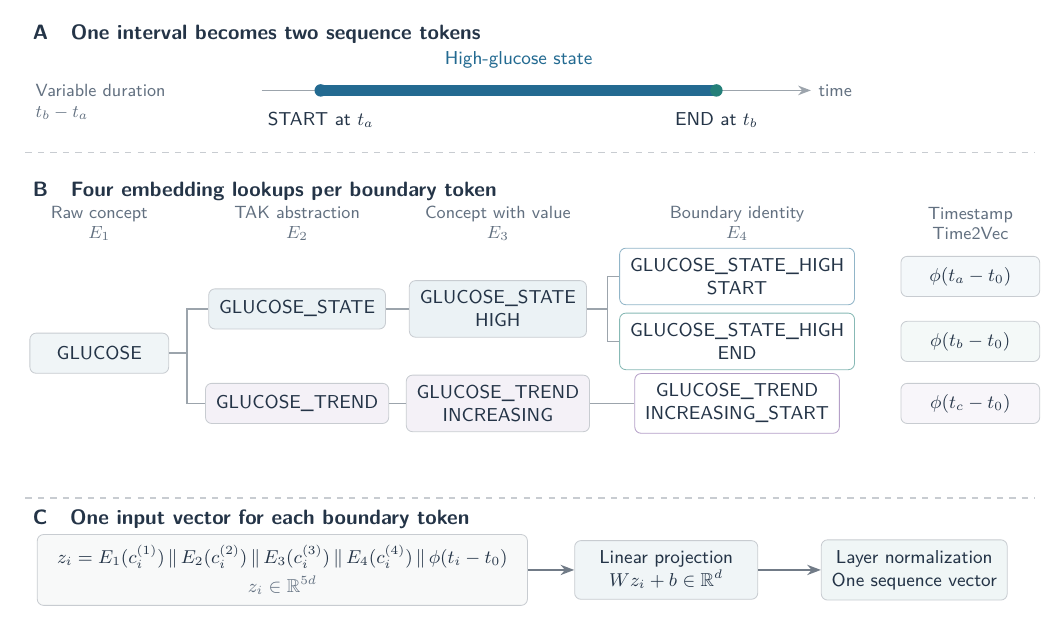}
  \caption{Hierarchical boundary-token embedding. (A) Each interval contributes START and END tokens. (B) Connections denote shared hierarchy entries: the state boundary pair shares $E_1$--$E_3$, while glucose state and trend share $E_1$. Here $t_0$ is admission time and $t_c$ the trend start. (C) The four embedding lookups and Time2Vec vector are concatenated and projected into one input vector per boundary token. The hierarchy adds no sequence positions.}
  \label{fig:hierarchy-tokenization}
\end{figure*}

\paragraph{Computing infrastructure.}
Both INTERVenE variants were trained on a single NVIDIA RTX~A5000 (24\,GB VRAM) using PyTorch 2.4.1+cu124, CUDA 12.4 (cuDNN 9.1.0), and Python 3.11. All experiments use a fixed random seed of 42. The full pipeline (all three phases) runs without multi-GPU or distributed training.

\paragraph{Optimizer and regularization.}
All phases use AdamW with weight decay $0.001$ and gradient clipping at $\mathrm{max\_norm}{=}1.0$ applied at every update step. Betas differ by phase: Phase~1 (embedder pre-training in both variants) and INTERVenE-Ar Phase~3 use AdamW defaults $(0.9,\,0.999)$; INTERVenE-Enc Phases~2--3 and INTERVenE-Ar Phase~2 use $(0.9,\,0.95)$. The Phase~1 scheduler is ReduceLROnPlateau (factor $0.5$, patience $4$, min LR $10^{-6}$); Phases~2 and~3 use OneCycleLR.

\paragraph{Batch size and learning rates.}
The physical batch size is 16 with gradient accumulation over 4 steps (effective batch size 64). Phase~1 and Phase~2 use a peak LR of $3{\times}10^{-4}$; Phase~3 trains the task head at $10^{-4}$ with the backbone running at a fixed fraction ($0.1{\times}$ for INTERVenE-Enc, $0.01{\times}$ for INTERVenE-Ar). LR was selected from the range $[10^{-3},\,10^{-5}]$ based on validation loss behavior.

\paragraph{Encoder time output.}
The time head outputs non-negative hours via softplus; prediction and target are standardized only when computing the training loss, so inference requires no inverse normalization.

\paragraph{Legality masking (INTERVenE-Ar Phase 2).}
\label{app:legality-masking}
At each position $i$, a dynamic boolean mask $\mathbf{m}_i \in {0,1}^{V}$ is computed from the accumulated sequence state and applied to both the logits and the BCE targets, ensuring the model never receives gradient signal for illegal continuations. Illegality is determined by three stateful rules: (i) an interval \texttt{END} token is illegal if its corresponding \texttt{START} has not been seen; (ii) an interval \texttt{START} token is illegal if the same base or a conflicting value of the same concept is already open; (iii) meal tokens must follow a fixed cyclic order. The mask is derived deterministically from the TAK interval grammar at every training step and is recomputed autoregressively during generation.

\paragraph{Soft-kernel horizons and Phase 2 label exposure.}
The terminal BCE window is $H_\text{term}{=}168$\,h; the outcome risk-head horizon is $H_\text{out}{=}48$\,h (positive signal fires within 48\,h of an outcome event). Per-class decay $\tau_k$ is initialized at 48\,h for outcome tokens and 12\,h for terminal and default LM tokens; the terminal $\tau$ is frozen after an early diagnostic showed unconstrained widening, while the outcome $\tau$ remains learnable in Phases~2 and is frozen in Phase~3. Phase~2 pretraining trains on full patient trajectories and therefore sees post-48\,h tokens, as the generative objective requires learning future sequence structure. Outcome labels (whether a complication occurs in hours 48--336) are introduced only in Phase~3; inference is seeded solely with the 0--48\,h input window, so no task-relevant label information is accessible during Phase~2 pretraining.

\paragraph{Auxiliary curriculum.}
Auxiliary activation and weighting are phase-specific and distinct from learning-rate warmup. The encoder Phase-2 schedule is specified below; the decoder's staged schedule is described in the main text.

\paragraph{Encoder Phase-2 targets and weights.}
\label{app:enc-phase2-loss}
Let $t_i$ denote hours since admission. The position target is $t_i/336$, with MSE averaged over non-padding positions. For a masked position $i$, let $d_i$ be the smallest absolute time gap to an unmasked, non-padding event more than two seconds away. The local target is $\min(d_i/24,5)$; its MSE includes only masked positions with an eligible neighbor. Excluding near-simultaneous events avoids trivial local-gap targets.

The supplied configuration uses four MLM-only epochs, followed by simultaneous activation of both temporal auxiliaries with no weight ramp. Each weight is calibrated once from the epoch-mean training losses at the end of warmup:
\begin{equation}
\lambda_a=\min\!\left(400,\frac{c_a\,\overline{\mathcal{L}}_{\mathrm{MLM}}}{\overline{\mathcal{L}}_a}\right),
\quad a\in\{\mathrm{pos},\mathrm{local}\},
\end{equation}
where $c_{\mathrm{pos}}=0.40$ and $c_{\mathrm{local}}=0.15$. Calibration is deferred if the auxiliary loss is at most $10^{-8}$; its weight remains zero until calibration succeeds. Calibrated weights remain fixed thereafter. The fractions set relative contributions at calibration, subject to the weight ceiling of 400, rather than imposing a per-batch cap throughout training.

\paragraph{Decoder Phase-2 losses and schedule.}
\label{app:ar-phase2-loss}
The token objective applies class-weighted focal BCE ($\gamma=0.5$) to temporal soft targets over legal classes at non-padding prediction steps. Positive-target and zero-target entries are averaged separately, with the latter weighted by $\operatorname{clip}(0.85N_+/N_-,0.03,0.30)$, where $N_+$ and $N_-$ are their respective counts (each bounded below by one).

For legal classes $A_i$, positive-target classes $S_i\subseteq A_i$, and token logits $z_{iv}$, the set loss is
\begin{equation}
\mathcal{L}_{\mathrm{setCE}}=\operatorname{mean}_{i:S_i\ne\varnothing}
\left[\log\sum_{v\in A_i}e^{z_{iv}}-\log\sum_{v\in S_i}e^{z_{iv}}\right].
\end{equation}
Let $u_i=t_i/336$ be normalized time and $b_i=\mathbf{1}[u_{i+1}-u_i>10^{-8}]$. The timing auxiliary is
\begin{equation}
\mathcal{L}_{\mathrm{time}}=\mathrm{BCE}(\hat b_i,b_i)
+\mathrm{MSE}_{i:b_i=1}(\hat u_{i+1},u_{i+1}),
\end{equation}
with both terms restricted to non-padding prediction steps. The gate distinguishes simultaneous from advancing timestamps; regression estimates the next normalized timestamp only for advancing steps. The terminal auxiliary is
\begin{equation}
\mathcal{L}_{\mathrm{ttt}}=\mathrm{MSE}_{i\in I_{\mathrm{term}}}
\big(\hat\delta_i,\log(1+\delta_i)\big),
\end{equation}
where $I_{\mathrm{term}}$ contains non-terminal, non-padding prediction positions with an available future terminal event and $\delta_i$ is the gap in hours. For each outcome, positive and negative positions have temporal outcome targets greater than zero and equal to zero, respectively. The ranking loss averages $\log(1+\exp(r_n-r_p))$ over positive--negative outcome-logit pairs, then over outcomes with both classes present; at most 256 positive and 512 negative positions per outcome are sampled per batch.

The supplied configuration uses four token-only epochs, then activates set cross-entropy, timing, and time-to-terminal losses together without ramps. Ranking unlocks after two validation checks without relative improvement of at least $10^{-3}$ and ramps linearly over three epochs. Each auxiliary's target weight is calibrated once as
\begin{equation}
\lambda_a^*=\min\!\left(10,\frac{c_a\,\overline{\mathcal{L}}_{\mathrm{token}}}{\overline{\mathcal{L}}_a}\right),
\end{equation}
using epoch-mean training losses immediately before activation, with $c_{\mathrm{CE}}=0.50$, $c_{\Delta t}=0.50$, $c_{\mathrm{ttt}}=0.60$, and $c_{\mathrm{rank}}=0.20$. Calibration is deferred while the raw auxiliary loss is at most $10^{-8}$. Weights remain zero before activation and calibration, and fixed after any ramp. These fractions govern calibration, not continuously enforced loss-contribution caps.

\paragraph{Early stopping and actual convergence.}
Each phase is capped at 100 epochs; training halts when the validation loss fails to improve by a minimum relative margin of $0.001$ for 10 consecutive epochs (patience searched over $5$--$15$). Actual convergence: both variants share the Phase~1 embedder, which ran 27 epochs (best checkpoint at epoch 22). INTERVenE-Enc Phase~2 ran 99 epochs (best at epoch 94); Phase~3 ran 49 epochs (best at epoch 44). INTERVenE-Ar Phase~2 ran 32 epochs (best at epoch 27); Phase~3 ran 70 epochs (best at epoch 65). The 2{,}000-resample patient-level bootstrap on the held-out test set serves as the per-metric uncertainty estimate in lieu of multi-seed retraining (see Limitations).

\paragraph{Generation horizon (INTERVenE-Ar).}
Autoregressive generation is bounded by two stopping criteria: a token budget of 2{,}000 and a temporal limit of 336 hours from admission. Generation also halts early upon emitting a terminal token (DEATH/RELEASE). In practice the temporal or terminal condition is binding; the token budget is a safety cap.

\section{Learned Input Embedding Geometry}
\label{app:embedding-geometry}

\begin{figure}[!htbp]
  \centering
  \includegraphics[width=\columnwidth]{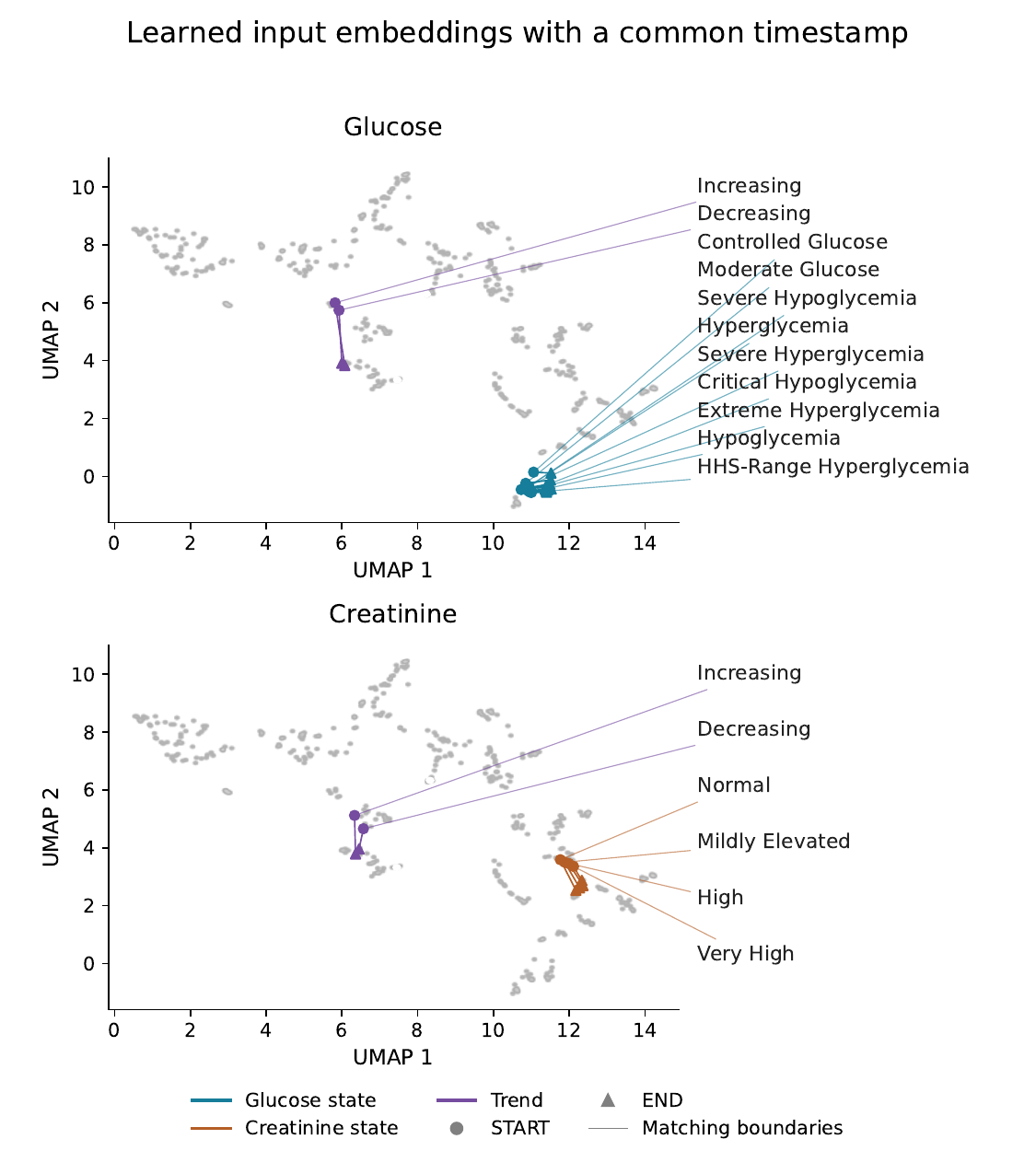}
  \caption{UMAP of trained input embeddings with timestamps held fixed. Glucose and serum-creatinine states and trends are highlighted against the boundary-token vocabulary (gray). Circles and triangles denote START and END tokens; connecting segments identify matching boundaries. Labels attach to START tokens. Original-space distances are summarized in the text.}
  \label{fig:embedding-geometry}
\end{figure}

We inspected the trained embedder from a saved INTERVenE-Enc Phase-3 checkpoint (epoch 38). Embeddings were extracted after projection and layer normalization, before the Transformer blocks, with a common admission timestamp and dropout disabled. Figure~\ref{fig:embedding-geometry} shows boundary-token embeddings using UMAP (cosine metric, 15 neighbors, minimum distance 0.15, seed 19), highlighting glucose and serum-creatinine concepts.

For named clinical state and trend values in the original embedding space, median cosine distances were 0.207 for matching START/END pairs, 0.220 for different values within the same state or trend concept, 0.305 for states and trends sharing a raw concept, and 0.442 for different raw concepts. This graded proximity was consistent with the hierarchy; it describes the trained representation without isolating the hierarchy's contribution from training.

\section{Sparse-Transcoder Interpretability}
\label{app:xai}

This appendix expands the exploratory attribution analysis summarized in the main text. We train one JumpReLU transcoder \citep{rajamanoharan2024jumping} per INTERVenE-Enc block on activations captured from a 1{,}500-patient stratified subset of the test cohort. Each transcoder reconstructs its block's MLP output from an overcomplete sparse dictionary (16$\times$ expansion; 2{,}048 features per layer). Reconstruction quality is high (per-layer $R^2=0.97$--$0.995$), while sparsity remains between approximately 5--19\% active features. The transcoders are used only for attribution and are never substituted into the deployed network. Instead, a gradient-decoupled hook preserves the forward computation exactly while routing gradients through the sparse feature basis, yielding activation$\times$gradient scores through the transcoder paths. These scores describe local model sensitivity; clinical names derive from the input tokens, rather than an established clinical interpretation of each latent feature.

\begin{figure*}[!t]
  \centering
  \includegraphics[width=\textwidth,trim=0 0 0 35bp,clip]{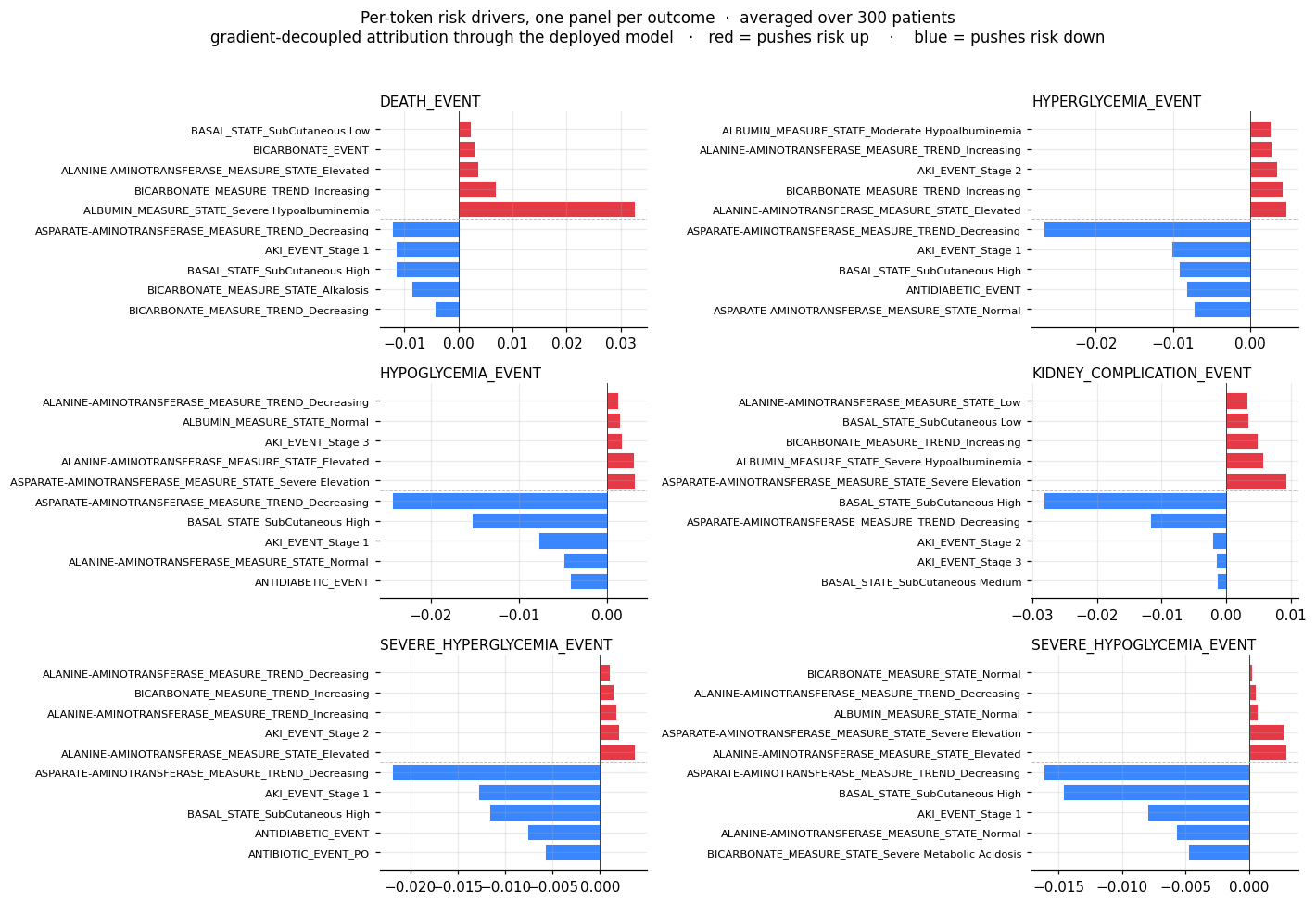}
  \caption{Per-token attribution scores for all six forecast outcomes, obtained from INTERVenE-Enc by gradient-decoupled sparse-transcoder attribution and averaged over 300 test patients. Within each panel the bars are signed gradient$\times$activation contributions summed over layers and features; red and blue denote positive and negative scores, respectively. All displayed tokens are named KBTA concepts (states, trends, and events). x-axis scales are panel-specific, so bar lengths should be compared only within a panel.}
  \label{fig:xai-all}
\end{figure*}

Attribution sums the signed gradient$\times$activation of an outcome's risk logit over all (layer, feature) pairs at each token, yielding the net contribution of each token to the predicted outcome, averaged here across 300 test patients. Interval \texttt{START}/\texttt{END} boundaries are collapsed so a single interval contributes one bar. Figure~\ref{fig:xai-all} reports the per-token drivers for all six outcomes (the main text presents the DEATH panel); across outcomes the drivers are clinically coherent and expressed entirely in named KBTA concepts.

\end{document}